# Informatics for Food Processing

Gordana Ispirova[1], Michael Sebek[2], Giulia Menichetti[1, 2, 3, *]

[1]Channing Division of Network Medicine, Department of Medicine, Brigham and Women's Hospital, Harvard Medical School, Boston, MA, USA
[2]Network Science Institute and Department of Physics, Northeastern University, Boston, MA, USA
[3]Harvard Data Science Initiative, Harvard University, Boston, MA, USA

## Abstract

This chapter explores the evolution, classification, and health implications of food processing, while emphasizing the transformative role of machine learning, artificial intelligence (AI), and data science in advancing food informatics. It begins with a historical overview and a critical review of traditional classification frameworks such as NOVA, Nutri-Score, and SIGA, highlighting their strengths and limitations, particularly the subjectivity and reproducibility challenges that hinder epidemiological research and public policy. To address these issues, the chapter presents novel computational approaches, including FoodProX, a random forest model trained on nutrient composition data to infer processing levels and generate a continuous FPro score. It also explores how large language models like BERT and BioBERT can semantically embed food descriptions and ingredient lists for predictive tasks, even in the presence of missing data. A key contribution of the chapter is a novel case study using the Open Food Facts database, showcasing how multimodal AI models can integrate structured and unstructured data to classify foods at scale, offering a new paradigm for food processing assessment in public health and research.

*To whom correspondence should be addressed: giulia.menichetti@channing.harvard.edu

## 12.2 Introduction to Food Processing

As concern grows over the health impacts of processed foods[1–3], researchers, policymakers, and consumers alike are asking a critical question: *What does it really mean for a food to be processed?* The answer is far from simple. While we increasingly rely on epidemiological data to draw connections between diet and disease, the upstream task of defining and classifying "processed food" remains complex and contested. Broadly, food processing refers to any alteration made to a raw agricultural product that affects its form, flavor, shelf life, or safety. This definition is recognized by major organizations such as the European Food Information Council (EUFIC), the United States Department of Agriculture (USDA), and the Food and Agricultural Organization (FAO) of the United Nations (UN)[4–6]. By this standard, nearly all food we consume is processed to some degree, from the boiling of vegetables at home to the industrial milling of grains into flour.

Yet not all processing is equal. Processing encompasses a wide design space, and its nutritional and health implications depend on the methods used and the purposes they serve. Cooking a tomato is not the same as engineering a shelf-stable, hyper-palatable snack. The spectrum of food transformation spans simple acts like chopping and heating to complex industrial formulations involving additives, emulsifiers, fermentation and extrusion technologies, each developed to achieve specific functional, sensory, or preservation goals. Understanding which processes provide nutritional benefits, which introduce potential risks, and under which circumstances is central to modern nutrition science. To appreciate the evolving relationship between food processing and human health, it is first necessary to examine how food preparation practices have developed over time and how technological and societal shifts have shaped what we eat.

Food processing, therefore, should not be interpreted as a single attribute but as a multidimensional concept reflecting different transformations, design choices, and technological purposes. Its potential health implications cannot be fully understood by examining individual foods in isolation, but rather by considering how processed foods contribute to overall dietary patterns and the nutritional structure of the diets in which they are embedded.

### 12.2.1 Historical Trends and Drivers

Food processing is one of humanity's oldest innovations. From roasting meat over fire to fermenting vegetables, early food transformation techniques were born out of necessity—for preservation, safety, and seasonal scarcity. For millennia, the methods and intensity of food processing were largely shaped by geography, class, technology, and labor. In ancient societies, elaborate techniques like milling or breadmaking were often reserved for elites; white bread, for instance, was a status symbol in the Roman Empire[4]. Preservation practices like salting, fermenting, and cooling date back at least 13,000 years and helped societies endure poor harvests and long winters[7,8]. These traditional methods persisted until the Industrial Revolution, which radically transformed both the purpose and scale of food processing. In the 19th century, inventions such as canning, pasteurization, oil hydrogenation, and refrigeration laid the groundwork for what would become the global, industrialized food system[9].

As factory work drew people into cities, diets shifted alongside infrastructure. Trains and steamboats enabled rapid distribution of food across vast distances, while electricity and crude oil unlocked faster production and storage. The invention of Freon in the mid-20th century made home refrigeration affordable and widespread, paving the way for ready-to-eat (RTE) meals that required little or no preparation. In their early forms, many of these meals were dry, shelf-stable products designed to last at room temperature. From the debut of RTE breakfast foods like Kellogg's Corn Flakes in 1906 to the success of Swanson's TV dinners in the 1950s[10], these innovations marked the beginning of a new era of convenience foods, tailored to demanding workdays and fast-paced living[11].

But convenience was only part of the story. As competition grew in the late 20th century, food manufacturers began to study consumer taste preferences scientifically[12]. The result was a new design principle: hyper-palatability, i.e., foods engineered to heighten pleasurable qualities like sweetness, saltiness, and richness by combining fat, salt, and sugar/carbohydrates at moderate to high levels[13]. These combinations, rarely found in natural foods, were crafted to maximize appeal and encourage repeat consumption[14].

Today, what we broadly define as highly processed foods (HPFs) — multi-ingredient, industrially formulated products that are typically ready to heat or eat and include components uncommon in traditional cooking — now dominate modern diets[15]. Their widespread availability, affordability, and engineered appeal have fundamentally reshaped what, how, and why we eat.

## 12.3 Current Frameworks for Evaluating and Quantifying Food Processing

Early attempts[16–18] to categorize food processing followed a straightforward model broadly capturing how many steps separate a raw agricultural product from its final form:

- Primary processing includes minimal interventions like cleaning, cutting, and refrigeration — measures that retain a food's original structure and composition.
- Secondary processing involves turning raw ingredients into more complex products through cooking, fermenting, or mixing, such as home-cooked or restaurant meals.
- Tertiary processing often refers to the industrial assembly of RTE meals using pre-cooked ingredients, additives, and packaging technologies. Additives, which are compounds or components added to food, serve various functions, such as extending shelf life, enhancing flavor, altering appearance, or restoring nutrients lost during processing[4,5]. For example, vitamin B1 may be reintroduced into refined grains, while paprika extract might be used to maintain visual appeal. More broadly, tertiary processing technologies are designed to: 1) preserve or delay food decay, 2) maintain or enhance quality and sensory attributes, 3) meet specific nutritional needs, and 4) reduce waste throughout the food supply chain[19]. These techniques require specialized equipment and resources, and are typically applied at the commercial scale, making them a key distinction between home-prepared meals and their industrially produced counterparts[6].

While this tiered system offers an intuitive framework, it quickly falls short under the complexity of today's global food supply. In practice, few foods reach our tables untouched by post-harvest modifications. For example, milk is pasteurized and fortified, grains are milled, and oils are refined and preserved. Even fresh fruits may be coated with wax to extend shelf life[20]. These processes, though common, blur the distinctions between processing categories. This blurring is particularly evident in secondary processing, where home-cooked meals often begin with ingredients that have already undergone extensive transformation.

To navigate this ambiguity, nutrition epidemiologists and food scientists have developed several classification systems and ontologies that describe foods according to their degree of processing. These systems were designed to translate the complexity of the modern food system into structured, actionable categories, particularly to support investigations into the health impacts of HPFs. In the following section, we examine some of the most widely used frameworks.

## 12.3.1 NOVA Classification

The NOVA classification system, first introduced in 2009 by Brazilian researchers led by Carlos Monteiro, organizes foods based on the extent and purpose of processing rather than on where or by whom the processing occurs[21]. Designed to support public health research and dietary surveillance, NOVA originally included three categories: minimally processed foods (MPFs), processed culinary ingredients (PCIs), and ultra-processed foods (UPFs). MPFs are foods that undergo minimal changes and retain their nutritional structure. They may be cleaned, portioned, frozen, boiled, dried, pasteurized, or packaged, but not substantially transformed. PCIs are substances derived from natural foods and intended for use in cooking rather than direct consumption. Examples include sugar, oil, flour, and salt, produced through processes such as milling, refining, and extraction. UPFs are industrially formulated foods primarily composed of PCIs and some MPFs. They are often pre-cooked and contain additives that enhance taste, appearance, and shelf life. These include RTE meals and branded, internationally distributed commercial food products, which require industrial equipment for production, such as hydrogenation and fortification. According to NOVA's developers, this class contributes to unhealthy eating patterns and increased risk of chronic disease. By 2016, the NOVA classification system included a fourth group of processed foods, creating NOVA as we know it today (Table 1)[22]. This group consists of foods made by combining MPFs with few PCIs, typically containing a few ingredients aimed at improving shelf life or altering sensory qualities. Some processes characteristic of processed foods include salting, smoking, canning, bottling, and non-alcoholic fermentation.

While the system offers an intuitive framework, its application is not without challenges. Commercial manufacturers rarely disclose their full production methods, making it difficult to assess the number and type of processing steps. As a workaround, analysts may employ deformulation, where ingredient lists are reverse-engineered to infer processing steps. This approach is time-consuming and imprecise due to the lack of exact ingredient amounts, leading to significant variability and self-regulation within the food industry[23,24]. Foods also consist of multiple components, each with different processing histories, further complicating how to weigh the ingredient processes for the overall food item.

Additives pose additional challenges. Indeed, chemical compounds like Allura Red AC (E129)[25] are clearly classified as artificial additives, but natural coloring agents like paprika extract raise questions about whether all additives used in industrial settings should be considered equal. Similarly, the addition of nutrients such as vitamin D in milk or vitamin B1 in flour complicates the assumption that all additives are markers of processing detrimental to human health.

| **Class** | **Label** | **Common Processes** | **Example Foods** |
|---|---|---|---|
| NOVA 1 | Unprocessed or minimally processed foods | Cleaning, portioning, freezing, boiling, drying, pasteurization | Apples, potatoes, salmon, chicken, beans, peanuts |
| NOVA 2 | Processed culinary ingredients | Milling, refining, extracting, purifying | Olive oil, table salt, butter, wheat flour |
| NOVA 3 | Processed Foods | Bottling, canning, non-alcoholic fermentation, salting, smoking | Wheat bread, cheese, beer, wine, jerky |
| NOVA 4 | Ultra-Processed Foods | Hydrogenation, interesterification, fortification | Ice cream, breakfast cereal, frozen pizza, instant noodles |

*Table 1: The four classes of the NOVA food classification system. Examples of common processes and foods found in each class are provided.*

Although many ambiguities remain, as the next section will explore, NOVA has become a widely adopted tool in nutritional epidemiology and public health for evaluating how food processing affects health outcomes.

## *Public Health Implications of NOVA*

NOVA has played a central role in investigating the health risks associated with UPF consumption. Between 2015 and 2019, 95% of studies on this topic relied on NOVA to categorize foods[26]. These studies have linked UPF consumption with increased risk of obesity[27], over-eating[1], type 2 diabetes[28], cardiovascular disease (CVD)[29], and other non-communicable conditions[30–35].

Despite this extensive body of epidemiological evidence, the underlying biological mechanisms driving these associations remain only partially understood. One proposed pathway involves the hyper-palatability of UPFs, which are often energy-dense and rich in salt, sugar, and fat. These sensory characteristics are believed to override satiety signals, promoting excess intake, even though the exact reward-related neural pathways remain an area of ongoing research[36]. This focus draws attention to the nutritional quality of UPFs, which is notably not a formal criterion within the NOVA classification system.

Hyper-palatability, however, is not simply a function of elevated levels of fat, sugar, and salt, nor can it be fully captured by statistical models that combine these nutrients linearly — a strategy researchers have attempted with mixed success[1,37,38]. Hyper-palatability also involves physical transformations to food that often alter the natural food matrix[39,40]. This matrix represents the complex physical and chemical organization of nutrients and bioactive compounds in whole foods and plays a crucial role in digestion, absorption, and satiety. UPFs frequently exhibit textures that promote rapid consumption, such as crunchiness or softness, low chew resistance, and melt-in-the-mouth consistency. These altered food matrices may compromise nutrient bioavailability, postprandial glycemic responses, and satiety levels[41–44].

The structural changes in the food matrix often correlate with the inclusion of industrial additives that further modulate how UPFs interact with our body. Additives such as non-nutritive sweeteners and emulsifiers have been linked to microbiota disruption, which in turn can trigger intestinal inflammation, impair glycemic control, and contribute to long-term metabolic dysfunction[45–47]. Artificial food dyes, though added purely for visual appeal, have also been shown in preclinical models to compromise gut barrier integrity and influence inflammatory responses[48,49].

These compositional and structural features of UPFs are compounded by a third, less visible layer of concern: chemical contaminants that are not disclosed on ingredient lists. UPFs are frequently exposed to neoformed compounds produced during high-temperature processing methods such as baking, frying, and roasting[50]. In addition, industrial chemicals like phthalates and bisphenols can leach into food from packaging materials and processing equipment[51,52]. Though not intentionally added as ingredients, these substances have been detected across a wide range of packaged foods and are increasingly recognized as contributors to metabolic dysfunction, reproductive and developmental abnormalities, and hormone-sensitive cancers[53–56]. These multifactorial and often invisible risks challenge conventional models of dietary assessment and raise critical questions about the long-term safety of industrialized food systems.

## *Scientific and Policy Critiques*

The NOVA system has profoundly reshaped how we study and discuss food processing. It has underpinned a new wave of epidemiological research, introduced the term "ultra-processed food" into policy discourse, and raised public awareness about the links between modern diets and long-term health.

Despite its widespread adoption and paradigm-shifting influence, NOVA remains a qualitative and descriptive tool, which has led to inconsistencies and ambiguity across studies. Foods are often classified differently depending on the assessor, as interpretations vary and NOVA's definitions have evolved, changing eight times between 2009 and 2017[57]. The reliance on subjective, labor-intensive assessments based on incomplete and heterogeneous data across studies and countries results in poor inter-rater reliability and limited reproducibility[58,59]. Studies have shown that trained experts frequently disagree on whether certain items should be considered ultra-processed, even when detailed ingredient lists are available. This subjectivity undermines both the repeatability and the scientific robustness of the classification.

NOVA has also faced critique from food scientists and industry experts who argue that it oversimplifies processing and fails to account for differences in nutritional quality and health risk across UPF subgroups[60]. For instance, nutrient-fortified whole-grain cereals and baked products, fermented foods like many yogurts, and plant-based alternatives are all classified as equally ultra-processed, despite evidence suggesting they may offer health benefits, including reduced cardiovascular risk[2,61]. Furthermore, NOVA does not differentiate between different types of processing, such as pasteurization or fermentation, which can preserve or enhance nutritional value, compared to other industrial techniques that may reduce nutrient density. This coarse-grained approach groups nutritionally diverse items under a single label, producing broad estimates of UPF prevalence — such as 60% globally[62], 70% in the U.S.[63] and Greece[64], and 80% in South Africa[65] — that lack the specificity needed to support targeted and actionable public health interventions. In some cases, the system's application has led to contradictory outcomes, where subcategories of foods included in the sustainable and traditional Mediterranean diet have been classified as 58.7% or 41.0% ultra-processed, respectively[64]. These examples underscore the need to refine and disaggregate the UPF category for more nuanced health assessment.

These limitations have constrained both the scope and precision of research into the health effects of UPFs, weakening public trust and limiting their integration into formal dietary guidelines[6]. One major challenge is the lack of a clear mechanistic understanding of how UPFs cause harm. Given the estimated high prevalence in global food supply, a blanket recommendation to avoid all UPFs risks oversimplifying the evidence. Such guidance may also exacerbate social and economic disparities, particularly for lower-income populations who often depend on affordable, shelf-stable food products[66]. Instead, there is a need for nuanced, mechanistically grounded guidelines that distinguish among types of UPFs based on both health impacts and socioeconomic realities. While observational studies consistently associate UPF consumption with adverse outcomes, the underlying causal mechanisms remain uncertain. Most evidence comes from observational designs, which are susceptible to confounding factors, including socioeconomic status and broader lifestyle variables. Only a few short-term randomized controlled trials exist, and these often struggle to isolate the effects of processing from other variables like energy density and nutrient composition[37]. Large-scale, long-term intervention trials, which would provide more definitive evidence, are currently lacking[67,68]. Although several interventional studies are underway, they primarily focus on intermediate outcomes and are not designed to assess long-term endpoints, such as chronic disease incidence or mortality[69].

Importantly, many of these limitations are not unique to NOVA. Other classification systems also suffer from low reproducibility and high subjectivity, stemming from reliance on expert judgment in the absence of standardized, structured data. This has led to growing calls within the scientific community for a more objective framework grounded in measurable biological mechanisms rather than variable interpretations[15]. Among the potential starting points, the nutritional profile of foods remains the only dimension that is consistently regulated and reported worldwide, and thus may serve as a practical foundation for systematizing food classification. Yet, without standardized, open-access, high-resolution data in nutrition science, interpretive disagreements will persist, with cascading effects on epidemiological and clinical findings. While overcoming the challenges of conducting large-scale clinical trials in nutrition remains difficult[69], progress can be made by improving the breadth and quality of underlying data. Strengthening the data foundations of future classification systems will better support experimental research and clinical studies, advancing nutrition science toward a more accurate, evidence-based, and data-driven discipline[70].

## 12.3.2 Other Food Processing Classification Systems

The launch of NOVA sparked greater interest in how food processing affects human health. As more evidence emerged about the effects of UPFs, researchers became increasingly eager to explore their components to uncover the biological mechanisms influencing health outcomes. This curiosity led to the development of various food processing classification systems intended to encompass the complexities of the global food supply. Consequently, while many classification systems exist, only a few, such as NOVA, are widely recognized in the public health community. Reflecting public interest in food processing's effects, these systems are typically tailored for either consumers or researchers.

Consumer-focused food classification systems emphasize simplicity and comprehension, ensuring they are quick and easy to use. These systems produce straightforward labels that help consumers make informed choices and their adoption is generally driven by two main approaches: 1) front-of-pack (FOP) labeling, such as Nutri-Score[71] and the Health Star Rating System[72], and 2) mobile applications like Yuka[73] and GoCoCo[74]. FOP labels usually require governmental backing for widespread adoption, making them mandatory for food manufacturers. This facilitates consumer interaction since every grocery store product features these labels. Meanwhile, mobile applications offer barcode scanners that enable consumers to quickly access a food product’s score online.

When designing food classification systems for research purposes, the primary goal is to enhance frameworks like NOVA to better clarify the relationship between food processing and non-communicable diseases[15,75]. These systems can be developed retrospectively, based on observed health outcomes in large cohorts, such as those from the European Prospective Investigation into Cancer and Nutrition (EPIC)[76], or by integrating additional variables into the classification process. For example, some systems go beyond additive content to account for overall ingredient composition and differentiate between beneficial and harmful types of processing. Notable efforts in this direction include studies from the University of North Carolina (UNC)[77] and the Système d’Information et de Gestion des Aliments (SIGA)[78]. Overall, research-oriented classifications aim to generate more refined groupings of food products to help uncover the mechanisms linking diet to disease. Next, we examine two widely used systems in public health, Nutri-Score and SIGA, in greater detail.

### *Nutri-Score*

As an example of consumer-oriented labeling systems, Nutri-Score[71] represents the most widely adopted front-of-pack (FOP) scheme currently used worldwide. Developed by French public health researchers, it is designed to offer a simplified, at-a-glance assessment of a product’s nutritional quality. Nutri-Score has been formally adopted by seven European countries—France, Spain, Belgium, Switzerland, Germany, Luxembourg, and the Netherlands—and is endorsed by public health authorities for its effectiveness in guiding consumers toward healthier choices[79,80]. Its widespread use has prompted many transnational food manufacturers to include Nutri-Score labels across products sold in the European Union.

Nutri-Score assigns foods a rating from A (dark green) for the most favorable options to E (dark orange) for the least. This score is calculated using a linear model based on a selection of seven nutritional and ingredient components per 100g of product: energy, sugar, saturated fat, sodium (negative factors, $N$, each awarded 0-10 points), and fiber, protein, and fruit/vegetable content (positive factors, $P$, each awarded 0-5 points). Each component contributes independently to the final score, and the formula is expressed as:

$$\text{Nutri-Score} = N - P. \tag{1}$$

The lower the summation of the individual components, the better the Nutri-Score rating and label the food receives (Table 2). The system uses separate scales for food, beverages, and fats/oils (Table 2), adjusting thresholds to account for differences in nutrient density.

| **Nutri-Score (Food)** | **Nutri-Score (Fats, Oils, Nuts)** | **Nutri-Score (Drink)** | **Nutri-Score Label** |
|---|---|---|---|
| ≤ 0 | ≤ -6 | Water | A |
| 1 to 2 | -5 to 2 | ≤ 2 | B |
| 3 to 10 | 3 to 10 | 3 to 6 | C |
| 11 to 18 | 11 to 18 | 7 to 9 | D |
| 19 to 40 | 19 to 40 | 10 to 40 | E |

*Table 2: Nutri-Score rating system ranges for foods, oils, and drinks at each Nutri-Score label.*[81]

Importantly, Nutri-Score does not account for the processing methods used to produce a food item and therefore does not distinguish between whole, minimally processed, and highly processed products. Instead, it relies solely on the three nutrients driving hyper-palatability (sugar, fats, and salt) as negative factors to help identify HPFs. Conversely, positive elements such as fruit, vegetable, fiber, and protein content can help identify MPFs.

A key limitation of Nutri-Score lies in the independent scoring of each component. For instance, 1 gram of salt per 100g contributes five points across all products, regardless of their broader context or formulation. This means that reformulated highly processed products, designed to lower sugar, fat, or salt content, can receive favorable scores, potentially misleading consumers about their overall health profile. At the same time, PCIs such as olive oil, which often have higher fat content per 100g but may offer health benefits when consumed in moderation, have been penalized by earlier versions of the algorithm. As a result, Nutri-Score has faced criticism for oversimplifying nutritional evaluation and failing to reflect the degree of processing. In response to these concerns, an updated Nutri-Score algorithm was released in 2022, refining the scoring thresholds and introducing a dedicated scale for fats, oils, nuts, and seeds to more accurately evaluate many PCIs[80]. Nevertheless, some experts continue to advocate for integrating additional criteria such as ingredient quality, presence of additives, and extent of industrial processing, to develop a more holistic and health-relevant evaluation of food products.

Regardless of these challenges, Nutri-Score remains one of the most widely used and scientifically validated labeling systems. Its key strengths lie in simplicity, transparency, and its proven ability to influence consumer behavior and encourage reformulation by food manufacturers. As discussions continue to refine its algorithm, Nutri-Score may evolve further to incorporate emerging evidence on food processing while continuing to support informed, health-conscious dietary choices.

### *SIGA*

All identified risks associated with food processing under NOVA fall into the NOVA 4 category, which encompasses a broad and diverse range of UPFs. Hence, classification systems such as SIGA seek to enhance the granularity of NOVA across all processing levels, with particular attention to finer distinctions within the heterogeneous UPF category[78]. SIGA presents an advanced framework for evaluating the levels of food processing, connecting two essential viewpoints: the holistic perspective, which examines the structural integrity and interactions within food components, and the reductionist perspective, which emphasizes individual ingredients and technological changes. This model is an evolution of the NOVA classification, offering a more detailed differentiation among food categories by considering further criteria such as the impact of processing on the food matrix, and by rigorously identifying industrial markers of ultra-processing (MUP) and their documented health risks. MUPs are derived from technological processes related to cracking or synthesis, and they can serve as either ingredients or additives in food products.

SIGA takes NOVA classes and further divides them into two or more categories, converging to a total of seven classes (Table 3). NOVA 1 is separated into unprocessed foods and those that undergo minimal processing, distinguishing between raw milk and pasteurized milk, raw eggs and bleached eggs, and raw fruits and waxed fruits. Importantly, SIGA assigns the PCIs of NOVA 2 based on their processing; for example, whole wheat flour is found in Class 1 while refined wheat flour is assigned to Class 2. NOVA 3 has been divided according to the concentration of salt, sugar, and fat within the food item, following the recommendations of the Food Standard Agency of the United Kingdom (FSA) and the World Health Organization (WHO). Foods containing less than 1.5 g/100g salt, 12.5 g/100g sugar, and 17.5 g/100g fat, or beverages with less than 0.75 g/100g salt, 6.25 g/100g sugar, and 8.75 g/100g fat are considered nutritionally balanced (Class 3), while foods or beverages exceeding at least one of the thresholds are classified as imbalanced (Class 4).

By including criteria such as fat, sugar, and salt content, SIGA can differentiate foods based on hyper-palatability, designating those that may encourage overeating as more processed within their SIGA group (Table 3, Column 2). SIGA further distinguishes UPFs by the presence of MUPs within the ingredient list. A single MUP in a food item is sufficient for the food to be classified as a UPF. Similar to the breakdown of NOVA 3, SIGA further categorizes NOVA 4 into three classes through nutritional balance assessment, presence of MUPs, and presence of MUPs with known or uncertain health risks, according to the European Food and Safety Authority (EFSA) and the French Agency for Food, Environmental and Occupational Health and Safety (ANSES). Ingredients like refined oils, starches, xanthan gum, and lecithin are MUPs with no known health risks, placing foods containing these ingredients in Class 5 or Class 6, depending on their nutritional balance. Foods with health-risk MUPs, such as hydrolyzed sugars, modified starches, and sodium nitrite, are categorized as Class 7.

A large-scale analysis of over 24,000 packaged food products from French supermarkets using the SIGA system found that approximately 67% were classified as ultra-processed, with 54% exhibiting multiple markers of ultra-processing[78]. Among products containing more than five ingredients, 75% were categorized as ultra-processed, underscoring the strong correlation between complex formulations and industrial food production. These items often contained MUPs like refined oils, hydrolyzed sugars, modified starches, and synthetic flavorings, aligning with public health concerns regarding high UPF consumption, as defined by NOVA.

| NOVA class | SIGA groups | SIGA class | Description | Examples |
|---|---|---|---|---|
| NOVA 1 | A0 | 1 | Unprocessed foods | Raw milk, raw eggs, raw nuts, raw meat |
| | A1, A2 | 2 | Minimally processed foods and culinary ingredients | Pasteurized milk, refined grains, 100% fruit juice, ground beef, honey, butter, salt, table sugar |
| NOVA 3 | B1 | 3 | Nutritionally balanced processed foods | Canned beans, bread, canned fish, hard cheeses, fruit purees |
| | B2 | 4 | High salt, sugar, and/or fat processed foods | sauces, pickles, salted nuts, feta cheese, smoked meats, shortbread, salted butter |
| NOVA 4 | C01 | 5 | Nutritionally balanced ultra-processed foods with at least 1 MUP | Milk substitutes, soups, ravioli, meats cooked in refined oils |
| | C02 | 6 | High salt, sugar, and/or fat ultra-processed foods with at least 1 MUP | Potato chips, fruit jams, fried meats, salted nuts roasted in refined oils |
| | C1 | 7 | Ultra-processed foods with at least 1 health risk MUP | Meat substitutes, ketchups, chocolate spreads, nuggets, ice creams, hydrogenated oils, margarine |

*Table 3:The SIGA classification system compared to the NOVA classification system.*

However, unlike NOVA, SIGA has not yet been widely adopted in epidemiological research. This reflects important differences in the underlying logic of the two systems. NOVA, developed by nutrition epidemiologists, was designed to classify model foods commonly used in food frequency questionnaires and 24-hour recalls — foods that often lack detailed ingredient-level data and primarily report nutrient content. SIGA, by contrast, was built for application to real-world branded food products and relies on ingredient lists and nutrition facts available on the food packaging to detect MUPs and evaluate nutritional balance. These contrasting approaches reveal a broader gap between epidemiological research methods and the complexities of the commercial food supply, which may limit the portability and interoperability of classification systems between domains.

### 12.3.3 Food Processing Ontologies and Indexing Systems

The food processing classification systems described so far were primarily developed within nutritional epidemiology to evaluate health implications related to processed food consumption. In contrast, food scientists have independently developed ontologies and indexing systems that capture a wider range of food attributes, encompassing not only processing but also composition, source, and preparation methods. An ontology is a structured vocabulary that defines standardized terms and their relationships, enabling consistent tagging, classification, and integration of information across datasets. In the food and nutrition domain, ontologies serve as hierarchical frameworks that organize complex food-related data and promote interoperability across platforms and studies. While no ontology was originally created to specifically classify food processing levels, several incorporate elements related to processing and can be adapted for relevant research. Notable examples include LanguaL, FoodEx2, and FoodOn. These systems help standardize food descriptions, link ingredients to preparation techniques, and provide structured foundations for AI-based food classification models. Despite their potential the practical use of these ontologies and indexing systems in machine learning applications remains constrained by the limited availability of large, annotated datasets specifically designed to train models.

#### *LanguaL for Food Processing*

The LanguaL (Langua Alimentaria) classification system[82] is a hierarchical food indexing thesaurus designed to standardize food descriptions across databases and research applications. It provides a multi-faceted approach to describing foods based on their composition, origin, and processing methods. Such granularity is critical for updating national food composition tables as well as for supporting cross-country data harmonization and assessments of public health outcomes. LanguaL has been used for harmonization and linkage of food composition tables with consumption databases, for example by describing and classifying traditional Italian composite dishes[83] — capturing multi-ingredient formulations and cooking methods. Furthermore, LanguaL was used in standardizing descriptors for ready-to-eat products to support nutritional labeling, dietary exposure assessments, and regulatory applications[84]. LanguaL has also been used for the harmonization of nutrient databases across European countries, as its standardized descriptors facilitate the comparison and integration of food composition data essential for multinational nutritional surveillance[85]. The EPIC nutrient database project used LanguaL as part of its strategy to standardize nutrient databases across several European countries[86], ensuring that epidemiological studies could rely on harmonized food composition data when assessing the impact of food processing on nutrient intake and subsequent health effects.

Among its various facets, several are particularly relevant to food processing classification as they describe cooking methods, preservation techniques, physical transformations, and industrial treatments. These facets enable a systematic categorization of food processing at distinct levels, making LanguaL a valuable tool for assessing food transformation across large datasets.
The following LanguaL facets provide key descriptors of food processing and modification:

1. **Facet F: Cooking Method**
   This facet categorizes food based on the cooking techniques applied, such as boiling, frying, steaming, roasting, baking, and microwaving. Cooking significantly alters food texture, nutrient composition, and digestibility, making it a key determinant of processing levels.

2. **Facet G: Preservation Method**
   Foods are classified based on the preservation techniques used to extend shelf life, including freezing, canning, dehydration, irradiation, and vacuum-sealing. These methods help maintain food stability but may also impact nutritional value and food structure.

3. **Facet H: Packing Medium**
   This facet describes the medium in which food is packed or stored, such as brine, oil, syrup, vinegar, or vacuum packaging. The choice of medium influences food stability, sensory attributes, and potential chemical interactions.

4. **Facet J: Treatment Applied**
   This facet captures industrial and mechanical treatments such as pasteurization, fermentation, hydrogenation, fortification, and enzymatic processing. These treatments often define whether a food remains minimally processed or transitions into a more refined or ultra-processed category.

5. **Facet K: Physical State, Shape or Form**
   Foods are described based on their physical state, such as powdered, granulated, shredded, liquid, or whole form. The degree of structural modification influences processing classification and ingredient functionality in food formulations.

6. **Facet M: Storage and Use Conditions**
   This facet indicates how a food product is stored or intended to be used, such as shelf-stable, refrigerated, frozen, or ready-to-eat. Storage requirements reflect processing intensity, as highly processed foods tend to have extended shelf lives due to stabilizers and preservatives.

Facets A, B, and Z primarily describe food sources and biological origins; however, some subclasses within these facets provide additional insights into food processing characteristics:

1. **Facet A: Product Type**
   This facet categorizes foods based on their general classification, distinguishing between raw, semi-processed, and processed food items. Subclasses include refined foods such as milled grains, processed dairy, and extracted oils, which reflect varying levels of industrial transformation.

2. **Facet B: Source of the Food Product**
   Although primarily focused on the biological source of foods, certain subclasses describe whether an ingredient has been modified or isolated from its original source (e.g., whole grain vs. refined grain, whole milk vs. powdered milk).

3. **Facet Z: Processing Technologies and Industrial Applications**
   Some subclasses within this facet relate to specific processing technologies and industry-defined food categories, such as fermentation techniques, enzymatic treatments, and specialized food formulations. These descriptors help track industrial processing methods across food supply chains.

In summary, the structured vocabulary provided by LanguaL enables researchers, policymakers, and AI-driven food classification models to systematically assess food processing levels across large datasets. Its hierarchical classification allows for possible automated food processing classification in machine learning applications. By leveraging facets related to cooking, preservation, treatment, and physical transformation, LanguaL can offer a granular view of food processing that enhances the accuracy of nutritional and epidemiological assessments.

### *FoodEx2*

Developed by EFSA, FoodEx2[87] is a standardized food classification system designed for food safety monitoring, dietary exposure assessment, and regulatory reporting. It provides a hierarchical structure that categorizes foods based on their composition, processing level, and intended consumption. Key aspects include:

- Categorization of raw, minimally processed, and processed foods.
- Tracking of food additives, contaminants, and exposure risks.
- Linking food descriptions to consumption surveys and exposure models.

FoodEx2 underpins EFSA's DietEx tool, which estimates chronic dietary exposure to chemicals by mapping consumption data from the Comprehensive European Food Consumption Database to FoodEx2 codes[88]. It also drives the ImproRisk model, enabling chronic exposure assessment to a wide range of food-borne hazards through FoodEx2–based categorization[89]. Beyond chemical risk, FoodEx2 was employed to harmonize the Italian IV SCAI children's survey data standardizing descriptions of foods, beverages, and supplements consumed by children between 2017 and 2020 in compliance with EFSA's EU Menu guidelines[90].

### *FoodOn Ontology*

FoodOn[91] is a comprehensive ontology that models food products, ingredients, and food transformation/processing techniques in a standardized structure. The ontology presents a hierarchical categorization of food products, ranging from raw agricultural commodities to processed foods. A key feature of FoodOn, especially relevant to food processing, is its "Food Transformation Process" class, which provides a structured vocabulary for describing how food products are altered throughout the production, processing, and distribution pipeline. This class is organized hierarchically, allowing food items to be classified according to their degree of processing, transformation method, and the technological interventions applied to them.

FoodOn categorizes food transformations into distinct process classes, including:

1. **Physical Transformations**
    - Mechanical processing: Grinding, milling, crushing, chopping, slicing.
    - Phase transitions: Dehydration, freezing, thawing, sublimation (freeze-drying).
    - Structural modifications: Homogenization, emulsification, extrusion.

2. **Thermal Processing**
    - Heat-based transformations: Boiling, steaming, roasting, frying, baking, grilling.
    - Cold processing: Refrigeration, freezing, deep-freezing.
    - Pasteurization and sterilization: High-pressure processing (HPP), ultra-high-temperature (UHT) treatment.

3. **Chemical and Biochemical Transformations**
    - Fermentation processes: Lactic acid fermentation (yogurt, kimchi), alcoholic fermentation (wine, beer).
    - Acidification and alkalization: Pickling, pH modification.
    - Enzymatic processes: Curing, proteolysis, hydrolysis.
    - Food fortification: Addition of vitamins, minerals, amino acids.
    - Addition of preservatives and additives: Emulsifiers, stabilizers, colorants, anti-caking agents.

4. **Combination and Formulation Processes**
    - Mixing and blending: Ingredient incorporation (e.g., salad dressings, spice blends).
    - Reconstitution: Powdered foods mixed with liquids (e.g., reconstituted fruit juice, instant soups).
    - Binding and texturization: Use of hydrocolloids, gelatinization, protein structuring.

5. **Packaging and Storage Processing**
    - Modified atmosphere packaging (MAP).
    - Vacuum sealing and controlled-environment storage.
    - Edible coatings for preservation.

Each transformation step within FoodOn is connected to both input and output food states, enabling traceability from raw ingredients to final processed products. This structured representation aids researchers in analyzing how food processing impacts nutritional quality, shelf life, and safety. FoodOn consolidates multiple food categorization systems, such as the USDA food composition databases[92], LanguaL[82], and the Ontology for Biomedical Investigations (OBI)[93], facilitating interoperability among scientific, industrial, and regulatory datasets. Each food classification ontology integrated offers a complementary evaluation of food processing:

- LanguaL (for food descriptions and processing methods).
- FoodEx2 (for regulatory and exposure assessment applications).
- Open Food Facts (for real-world packaged food processing analysis).
- OBI and ENVO (Environment Ontology) (for linking food transformations with environmental factors).

By integrating these systems, FoodOn offers a unified ontology that links food composition, processing transformations, and health-related metadata.

## 12.4 Emerging Trends in Food Processing Assessment: From Qualitative Systems to AI-Driven Methodologies

As mentioned previously, recent studies[58,94] have shown that descriptive and subjective classification systems, such as NOVA, suffer from poor inter-rater reliability and reproducibility (see *Scientific and Policy Critiques*). These issues are further exacerbated by the inconsistent availability of data across food databases to match the descriptions. Meanwhile, more algorithmic systems like Nutri-Score and SIGA have attempted to introduce quantitative rigor into the assessment of product healthfulness by applying fixed thresholds to a few pre-selected nutrients. Yet, this approach brings its own challenges. First, fixed thresholds applied uniformly across food categories can penalize naturally nutrient-dense foods, while rewarding reformulated HPFs that are engineered to meet specific criteria without improving overall quality. Second, the expert selection of only a few nutrients, such as sugar, fat, and sodium, provides an informative but ultimately reductionist picture of food composition and processing. Processing does not just affect isolated biomarkers but alters the concentrations of multiple nutrients in a coordinated manner, with combinations that correlate strongly with the degree of processing[94]. Focusing on a small set of indicators makes these systems easy to manipulate through targeted reformulation, while ignoring important changes in the remaining food matrix. Third, scoring each nutrient independently using fixed cutoffs or expert-defined weights assumes linear relationships and fails to capture interactions between components, again encouraging optimization of individual nutrients at the expense of holistic food quality.

These limitations of one-size-fits-all nutrient scoring systems highlight the broader challenges faced by standard classification methods. By relying on predefined thresholds and expert-driven criteria, such systems often fail to capture the full complexity of food processing and are prone to subjective biases. This complexity instead calls for machine learning (ML) approaches, which are well-suited to detect the combinatorial changes in nutrient composition introduced by food processing. Data-driven models can empirically uncover reproducible patterns, refine classification schemes, and mitigate the inter-rater variability and selection bias that limit expert-based systems — all while offering far greater scalability. In fields such as genomics, ML has already transformed the landscape by enabling the discovery of complex, non-linear associations across massive datasets. Nutrition science may now be approaching a similar inflection point.

However, applying AI models to nutrition faces substantial obstacles. First, the underlying data is often incomplete or inconsistently structured. Real-world branded food products, such as those found in grocery stores, are typically described using nutrition facts and ingredient lists[66,95]. These formats differ markedly from the "model foods" used in epidemiological research, which represent composite averages and often include a broader range of nutrient values but lack ingredient specificity and real-market granularity[96]. This disconnect creates a major portability gap between public health tools developed in controlled research settings and their application to actual consumer environments.

Compounding this challenge is the limited transparency in commercial food manufacturing. Proprietary formulations and undisclosed processing methods often leave researchers with only the nutrient facts and ingredient list to work from, which provide limited insight into food production and allow for significant uncaptured variability[97]. Furthermore, AI models require large, well-annotated datasets with consistent class labels to perform effectively. In food processing informatics, both the scale and clarity of available data are often lacking. AI systems tend to thrive in scientific domains where open-access databases with rich, standardized metadata support cross-validation, benchmarking, and transferability. In contrast, nutrition science is constrained by a shortage of food composition databases that are comprehensively labeled according to standardized processing schemes, limiting the design and application of AI models. Even when classification or ontology annotations are available, they may introduce additional ambiguity, such as assigning univocal labels to only a small portion of the food supply or classifying the same item in multiple overlapping categories. These inconsistencies complicate supervised learning tasks and increase the risk that AI models may end up replicating human biases rather than revealing mechanistic truths, inheriting the same subjectivity they aim to replace.

Nonetheless, promising examples have begun to emerge. In the next section, we examine FoodProX[94], a machine learning model designed to estimate the degree of food processing based on nutrient composition. Despite current data limitations, it demonstrates the feasibility of developing classification systems that are both reproducible and adaptive, paving the way for a more evidence-based, data-rich future in nutrition science.

## 12.4.1 FoodProX

FoodProX[94] is a ML–based classifier designed to predict the degree of food processing by evaluating the full nutrient composition of a product. Unlike systems such as Nutri-Score or SIGA, which rely on expert-defined thresholds for a small subset of nutrients, FoodProX learns classification boundaries directly from data. Specifically, it maps nutrient concentration patterns to NOVA categories, allowing the model to detect systemic changes introduced by processing. This approach moves beyond hyper-palatability and nutrient-based scoring limited in scope by grounding decision-making in the broader physiological and chemical analysis of nutrient distributions. Indeed, many staple foods originate from once-living organisms, whose nutrient concentrations are shaped by complex metabolic networks[98] and constrained by physical and chemical properties[99]. Therefore, these concentrations tend to follow predictable, universal distributions across the food supply, well approximated by log-normals with a consistent logarithmic standard deviation at different levels of average concentrations (Figure 1a)[100,101].

As the foundational concept of hyper-palatability suggests, food processing systematically shifts these physiological ranges — either by removing or enriching certain nutrients or by introducing novel compounds — which can distort the biochemical balance that underpins human homeostasis[101]. Such coordinated deviations across multiple nutrients can be identified by ML models and linked to varying degrees of processing. For instance, chemical and physical transformations during food production, such as vitamin loss from milling or sodium addition for preservation, result in detectable compositional shifts. Importantly, these shifts need to be captured holistically. A clear example is the comparison between raw onions and fried, battered onion rings: in this case, approximately 75% of nutrients change in concentration by more than 10%, with more than half of the nutrients experiencing tenfold changes (Figures 1b–c).

In its original implementation, the FoodProX algorithm was developed as a multi-class random forest classifier trained on nutrient quantities (in grams per 100 grams of food) to estimate the probability that a food belongs to one of the NOVA categories. Its design reflected the data landscape available in 2019, particularly the limited availability of datasets annotated with NOVA labels. At the time, the main resource with NOVA annotations was the USDA's Food and Nutrient Database for Dietary Studies[106] (FNDDS) 2009–2010, the first notable application of the NOVA classification to epidemiological studies in the U.S.[102]. FNDDS is well suited to ML applications: it provides a rich nutrient profile for thousands of model foods, reporting between 65 and 102 nutrients depending on the cycle, with no missing values. However, only 2,484 items (34.25%) in this dataset were assigned a unique NOVA class, while the remainder were either left unclassified or required decomposition into their constituent ingredients.

These aspects contribute to the reproducibility and portability issues documented for NOVA classification. First, manual annotators working outside of FNDDS often rely on food composition databases that report model foods with varying levels of detail, primarily focusing on nutrient composition and lacking ingredient lists[103]. This limitation prevents ingredient decomposition and renders the assignment of NOVA labels highly subjective. Second, the ingredient fields in FNDDS were primarily designed to map model foods to other USDA datasets, such as the Standard Reference Database, for nutrient calculations. As a result, these ingredient descriptions lack the level of detail found in branded product ingredient lists and do not provide sufficient granularity to assess processing markers like food additives.

Leveraging the extensive nutrient panel available in FNDDS, three versions of the FoodProX model were trained using progressively smaller sets of nutrients (99, 62, and 12 variables) to simulate the transition from high-resolution food composition data for epidemiological studies to the more limited FDA-mandated nutrition facts found on branded food products[95]. Each model was trained using five-fold stratified cross-validation to ensure robust performance estimates across class distributions. Despite the reduction in input features, all models demonstrated commendable predictive ability and stability. The area under the receiver operating characteristic curve (AUC) remained consistently high across the nutrient panels. Even the 12-nutrient model exhibited $0.9804 \pm 0.0012$ for NOVA 1, $0.9632 \pm 0.0024$ for NOVA 2, $0.9696 \pm 0.0018$ for NOVA 3, and $0.9789 \pm 0.0015$ for NOVA 4. Additionally, the area under the precision-recall curve (AUP), which is particularly important for imbalanced classification tasks, also showed strong performance in the 12-nutrient model: $0.8882 \pm 0.0048$ for NOVA 1, $0.7468 \pm 0.0085$ for NOVA 2, $0.8723 \pm 0.0060$ for NOVA 3, and $0.9913 \pm 0.0006$ for NOVA 4. All metrics significantly exceed random baseline performance, underscoring the strong predictive signal embedded in nutrient composition. This is particularly notable given that NOVA's manual classification criteria do not formally incorporate nutrient content. The result highlights the latent structure within nutrient profiles that corresponds to the degree of processing, even without explicit expert labeling based on these features.

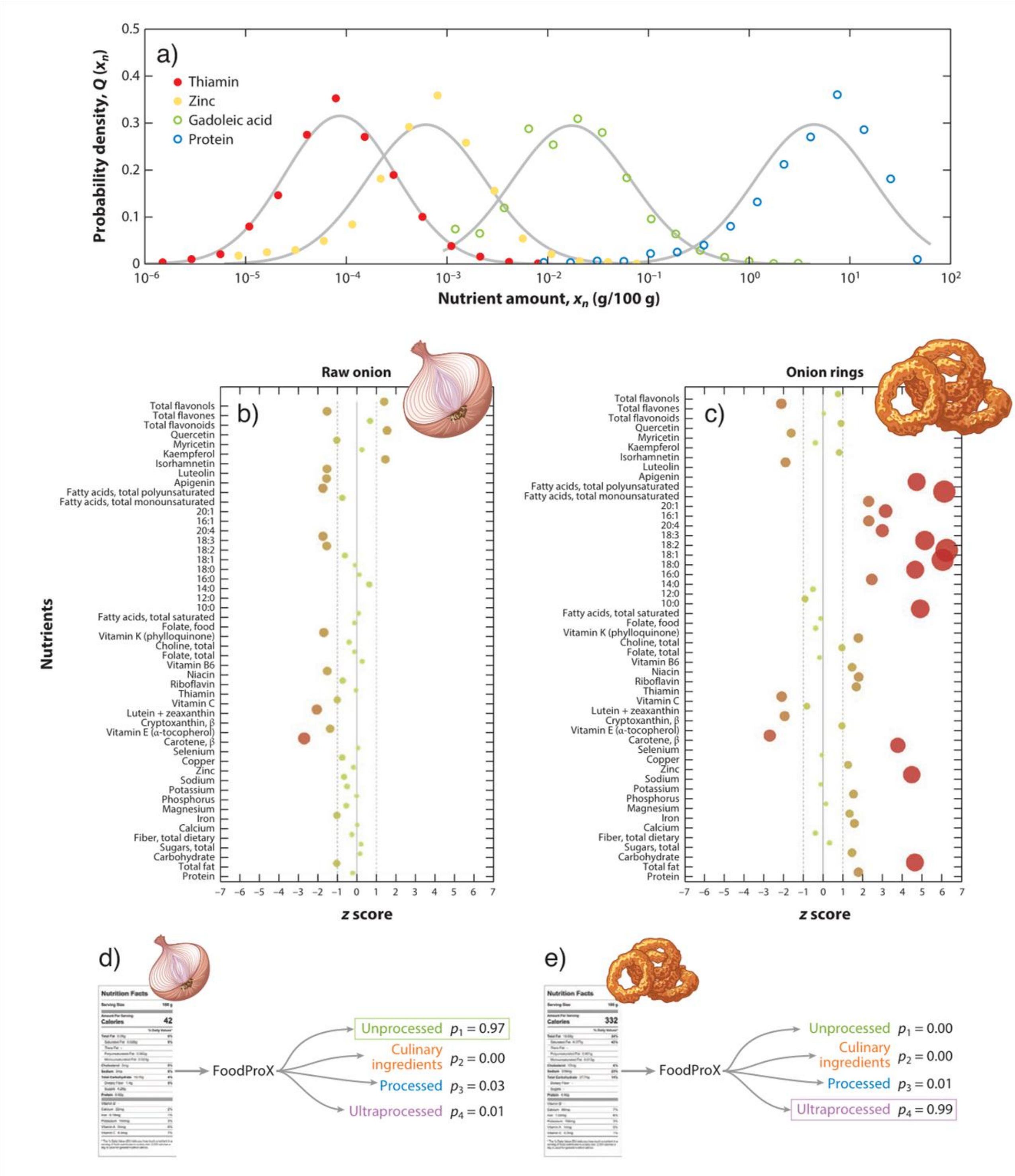

*Figure 1: **Large-scale analysis of nutrient concentrations in food. (a)** The concentration probability distribution for four nutrients across the 4,889 foods reported in NHANES 2009–2010 data, shown on a logarithmic horizontal axis. The four distributions are approximately symmetric on a log scale and have similar width and shape that are independent of the average concentration of the respective nutrient. Each symbol represents a histogram bin. **(b,c)** The observed common scale of nutrient fluctuations observed in the log space allows us to rescale all nutrients and compare them on a single plot, suggesting a methodology to detect foods with outlier concentrations. The pattern of nutrient outliers in different foods (quantified by a z-score in the log space) is informative of the type and extent of processing, as shown here for (b) 100g of raw onion compared with (c) 100 g of onion rings. **(d,e)** FoodProX is a random forest classifier that was trained over the nutrient concentrations within 100g of each food, tasking the classifier to predict its processing level according to NOVA. FoodProX represents each food by a vector of probabilities {$p_i$}, capturing the likelihood of the food being classified as an unprocessed food (NOVA 1), a processed culinary ingredient (NOVA 2), a processed food (NOVA 3), or an ultra-processed food (NOVA 4). The final classification label, highlighted with a box on the right, is determined by the highest probability. The probability values were rounded to two decimal places. Abbreviation: NHANES, National Health and Nutrition Examination Survey. [Reproduced with permission from Reference 101.]*

The strong performance of nutrient-based models is particularly encouraging given the current challenges in data standardization within nutrition science and the disconnect between epidemiological studies and real-world grocery environments. Most classification systems are designed for epidemiological research, yet their true public health impact depends on applicability at the consumer level, especially in grocery stores where people make daily food choices. Although both nutrition facts and ingredient lists contribute to the characterization of branded food products, the global inconsistency and poor regulation of ingredient data present significant barriers[66]. For instance, a GS1 UK data audit found an average of 80% inconsistency in product information across brands[104]. As a result, nutrition facts offer the most practical means of bridging the gap between curated model foods and real-world branded products, due to their consistent formatting, broad availability, and high reproducibility. In the European Union, for example, mandatory labeling includes total fat, saturated fat, carbohydrates, sugars, protein, salt, and caloric value, with member states often requiring additional fields[105]. These data are typically centralized into publicly accessible national databases, making them valuable assets for developing scalable food processing classification systems. The FAO's International Network of Food Data Systems (INFOODS) provides a global directory of such resources, supporting data interoperability across countries[106].

FoodProX assigns a set of probability scores $\{p_i\}$ to each NOVA class, with the final classification determined by the class with the highest probability (Figure 1d–e). These scores represent a point on the probability simplex, a set of four non-negative values that sum to one. Although this simplex exists in a four-dimensional space, the normalization constraint reduces it to a three-dimensional object geometrically equivalent to a tetrahedron. This structure enables diagnostic insights into model behavior. As illustrated in Figure 2a, manually labeled foods cluster distinctly near the corners of the simplex, indicating high classification confidence and justifying FoodProX's strong predictive performance.

Beyond the cross-validation dataset, FoodProX was used to classify the remaining FNDDS entries that lacked NOVA labels or needed further ingredient decomposition. This extended analysis predicts that 73.35% of the U.S. food supply, as represented in FNDDS, is ultra-processed — a figure confirmed by other estimates[63,107]. However, the core insight lies not in this percentage, but in the nature of the classifier's confidence. As shown in Figure 2b, these newly classified foods occupy the interior of the probability space rather than its corners, indicating more diffuse probability distributions and lower classification certainty. This pattern reflects the inherent ambiguity found in complex or composite foods, such as mixed dishes, where nutrient profiles do not align clearly with any single NOVA class. Rather than a limitation, this ambiguity is a key strength of FoodProX. It mirrors the challenges faced by manual assessors and highlights the limitations of hard classification in a domain where many foods exist along a continuum of processing. The probabilistic output of FoodProX thus provides a more nuanced, information-rich alternative to binary or categorical labeling, better capturing the diverse nutritional and processing characteristics within the food supply.

The limitations of discrete categorization in food classification motivated the development of FPro, a continuous scoring system that quantifies the degree of food processing along a gradient. While FPro builds upon the NOVA framework, especially in the absence of detailed compound-level data related to structural changes like cellular wall breakdown or specific industrial techniques, it offers a more nuanced alternative. By leveraging the full nutrient profile of a food item, FPro provides a ranking that is less prone to errors than rigid, binary classifications and supports more consistent comparisons across food items otherwise considered identically processed. Formally, FPro is defined as the orthogonal projection of a food's class probability vector $\{p_i^k\}$ onto the line within the probability simplex that extends from the minimally processed vertex (1,0,0,0) to the ultra-processed vertex (0,0,0,1). The score for item $k$ is given by:

$$FPro_k = \frac{1-p_1^k+p_4^k}{2}. \quad (2)$$

This formula captures the trade-off between the FoodProX model's confidence in classifying food item $k$ as NOVA 1 ( $p_1^k$) versus NOVA 4 ( $p_4^k$ ), the two endpoints of the processing spectrum (Figure 2b). As an example, the score is progressively higher for onion products as they undergo increasing levels of processing, ranging from approximately zero for raw ingredients (FPro = 0.0203 for raw onion) to one for UPFs (FPro = 0.9955 for onion rings) (Figure 2c).

FPro captures nuanced gradations in food processing levels by evaluating the nutrient composition of a product in its entirety, rather than assessing individual nutrients in isolation. Unlike traditional scoring systems that treat each nutrient independently, FPro is inherently non-linear: the impact of a single nutrient on the score depends on its interaction with all other nutrients in the food. This means that the same change in a nutrient's concentration can result in different shifts in FPro depending on the broader nutrient context. By learning from patterns of correlated nutrient variations within a fixed mass (100 grams), FPro estimates the likelihood that a food's overall nutrient profile resembles that of unprocessed or ultra-processed foods. For instance, although fortified products may contain similar levels of vitamins and minerals as whole foods, the algorithm can detect atypical concentration patterns, signatures indicative of industrial formulation, which contribute to a higher FPro score.

FPro's ability to translate complex nutritional profiles into a smooth numerical scale makes it ideally suited for recommendation systems that can guide targeted dietary interventions. For consumers, this ranking facilitates informed choices by highlighting less processed alternatives within familiar product categories such as cereals or cookies (Figure 2d). Recently, FPro has been

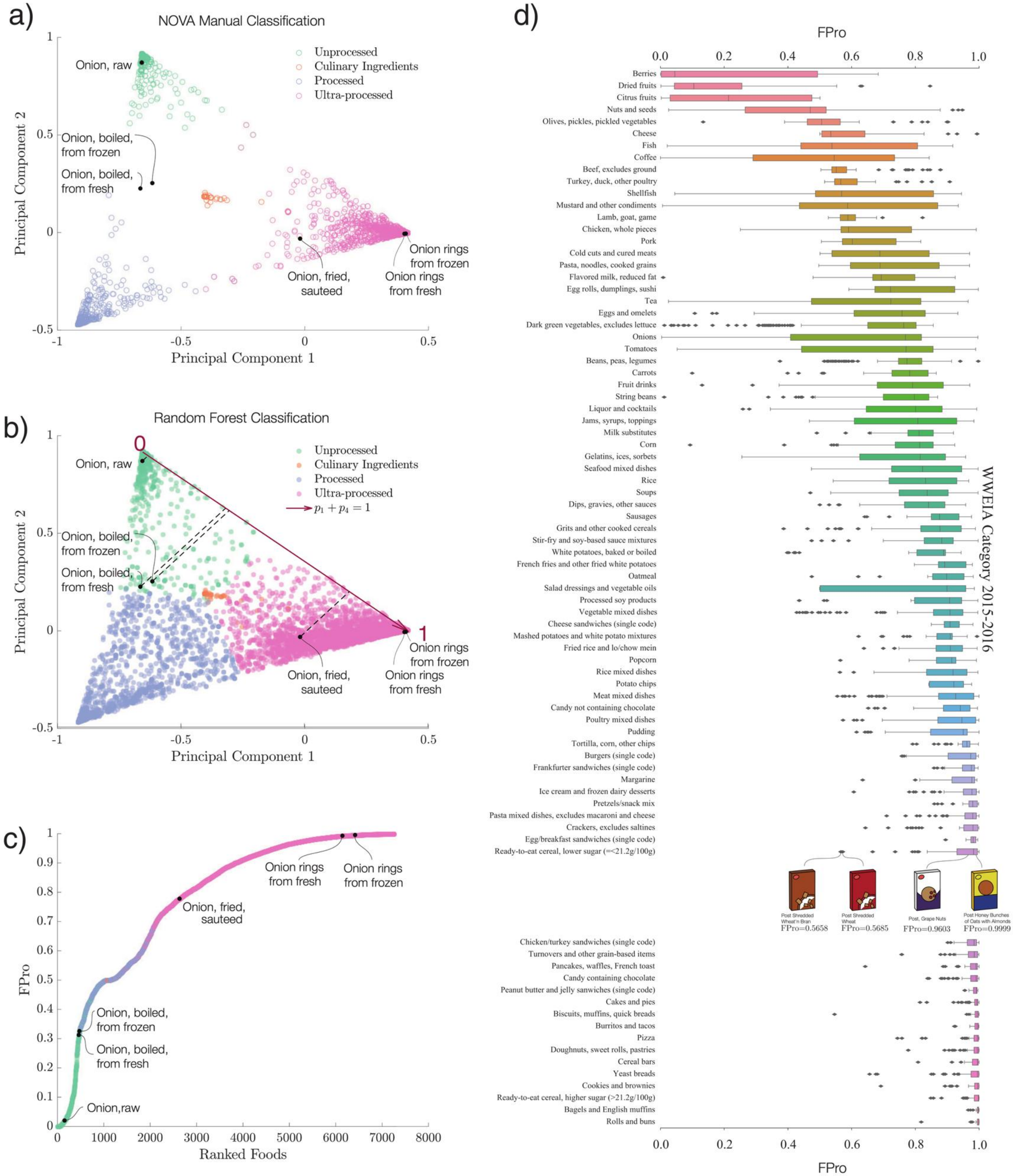
a)
NOVA Manual Classification
Unprocessed
Culinary Ingredients
Processed
Ultra-processed
Onion, raw
Onion, boiled, from frozen
Onion, boiled, from fresh
Onion, fried, sauteed
Onion rings from frozen
Onion rings from fresh
Principal Component 1
Principal Component 2
b)
Random Forest Classification
0
1
$p_1 + p_4 = 1$
Unprocessed
Culinary Ingredients
Processed
Ultra-processed
Onion, raw
Onion, boiled, from frozen
Onion, boiled, from fresh
Onion, fried, sauteed
Onion rings from frozen
Onion rings from fresh
Principal Component 1
Principal Component 2
c)
FPro
Ranked Foods
Onion rings from fresh
Onion rings from frozen
Onion, fried, sauteed
Onion, boiled, from frozen
Onion, boiled, from fresh
Onion,raw
d)
FPro
WWEIA Category 2015-2016
Berries
Dried fruits
Citrus fruits
Nuts and seeds
Olives, pickles, pickled vegetables
Cheese
Fish
Coffee
Beef, excludes ground
Turkey, duck, other poultry
Shellfish
Mustard and other condiments
Lamb, goat, game
Chicken, whole pieces
Pork
Cold cuts and cured meats
Pasta, noodles, cooked grains
Flavored milk, reduced fat
Egg rolls, dumplings, sushi
Tea
Eggs and omelets
Dark green vegetables, excludes lettuce
Onions
Tomatoes
Beans, peas, legumes
Carrots
Fruit drinks
String beans
Liquor and cocktails
Jams, syrups, toppings
Milk substitutes
Corn
Gelatins, ices, sorbets
Seafood mixed dishes
Rice
Soups
Dips, gravies, other sauces
Sausages
Grits and other cooked cereals
Stir-fry and soy-based sauce mixtures
White potatoes, baked or boiled
French fries and other fried white potatoes
Oatmeal
Salad dressings and vegetable oils
Processed soy products
Vegetable mixed dishes
Cheese sandwiches (single code)
Mashed potatoes and white potato mixtures
Fried rice and lo/chow mein
Popcorn
Rice mixed dishes
Potato chips
Meat mixed dishes
Candy not containing chocolate
Poultry mixed dishes
Pudding
Tortilla, corn, other chips
Burgers (single code)
Frankfurter sandwiches (single code)
Margarine
Ice cream and frozen dairy desserts
Pretzels/snack mix
Pasta mixed dishes, excludes macaroni and cheese
Crackers, excludes saltines
Egg/breakfast sandwiches (single code)
Ready-to-eat cereal, lower sugar (=<21.2g/100g)
Post Shredded Wheat'n Bran
FPro=0.5658
Post Shredded Wheat
FPro=0.5685
Post, Grape Nuts
FPro=0.9603
Post Honey Bunches of Oats with Almonds
FPro=0.9999
Chicken/turkey sandwiches (single code)
Turnovers and other grain-based items
Pancakes, waffles, French toast
Candy containing chocolate
Peanut butter and jelly sanwiches (single code)
Cakes and pies
Biscuits, muffins, quick breads
Burritos and tacos
Pizza
Doughnuts, sweet rolls, pastries
Cereal bars
Yeast breads
Cookies and brownies
Ready-to-eat cereal, higher sugar (>21.2g/100g)
Bagels and English muffins
Rolls and buns
FPro

*Figure 2: **(a)** Visualization of the decision space of FoodProX via principal component analysis of the probabilities $\{p_i\}$. The manual 4-level NOVA classification assigns unique labels to only 34.25% of the foods listed in FNDDS 2009–2010 (empty circles). The classification of the remaining foods remains unknown or must be further decomposed into ingredients. The list of foods manually classified by NOVA is largely limited to the three corners of the phase space, foods to which the classifier assigns dominating probabilities. **(b)** FoodProX assigned NOVA labels to all foods in FNDDS 2009–2010. The symbols at the boundary regions indicate that for these foods the algorithm's confidence in the classification is not high, hence a 4-class classification does not capture the degree of processing characterizing that food. For each food k, the processing score $FPro_k$ represents the orthogonal projection (black dashed lines) of $\vec{p}^k = (p_1^k, p_2^k, p_3^k, p_4^k)$ onto the line $p_1 + p_4 = 1$ (highlighted in dark red). **(c)** We ranked all foods in FNDDS 2009/2010 according to FPro. The measure sorts onion products in increasing order of processing, from "Onion, Raw", to "Onion rings, from frozen". **(d)** Distribution of FPro for a selection of the 155 Food Categories in What We Eat in America (WWEIA) 2015–2016 with at least 20 items. WWEIA categories group together foods and beverages with similar usage and nutrient content in the US food supply.[119] Sample sizes vary from a minimum of 21 data points for "Citrus fruits" to a maximum of 340 data points for "Fish". For each box in the box plots, the minimum indicates the lower quartile, the central line represents the median, and the maximum corresponds to the upper quartile. The upper and lower whiskers represent data outside of the inter-quartile range. All categories are ranked in increasing order of median FPro, indicating that within each food group, we have remarkable variability in FPro, confirming the presence of different degrees of processing. We illustrate this through four ready-to-eat cereals, all manually classified as NOVA 4, yet with rather different FPro. While the differences in the nutrient content of Post Shredded Wheat'n Bran (FPro = 0.5658) and Post Shredded Wheat (FPro = 0.5685) are minimal, with lower fiber content for the latter, the fortification with vitamins, minerals, and the addition of sugar, significantly increases the processing of Post Grape-Nuts (FPro = 0.9603), and the further addition of fats results in an even higher processing score for Post Honey Bunches of Oats with Almonds (FPro = 0.9999), showing how FPro ranks the progressive changes in nutrient content.[Reproduced with permission from Reference 94.]*

systematically applied to large datasets from major U.S. grocery stores, demonstrating its scalability and practical utility in real-world settings[66]. Moreover, the continuous nature of the score has proven especially effective for revealing broader trends, such as the correlation between processing level and price per calorie, a relationship that varies significantly across food categories.

Unlike expert-driven classification systems, FPro is a quantitative algorithm that leverages standardized inputs to generate reproducible, continuous scores, avoiding the use of arbitrary thresholds and maximizing discriminatory power across foods. Thus, FPro offers a more nuanced dietary guidance than broad classifications systems. For example, while both Post Shredded Wheat (FPro = 0.5658) and Post Grape-Nuts (FPro = 0.9603) fall into the same NOVA 4 category, FPro's continuous scale distinguishes between them, reflecting their differing degrees of processing (Figure 2d). This design supports sensitivity analyses and uncertainty quantification, both of which are typically absent in traditional frameworks. FoodProX, therefore, marks a significant step toward improving the objectivity and reproducibility of food classification through ML. Looking ahead, as the field's understanding of food processing deepens, the availability of increasingly large, heterogeneous, and unstructured datasets will demand more advanced modeling approaches. This shift is already paving the way for the adoption of deep learning architectures, including Large Language Models (LLMs), which are now expanding beyond natural language to capture complex biological and nutritional data.

## 12.4.2 Towards a Deeper Understanding of Food Processing

LLMs have emerged as powerful tools for extracting meaningful representations from textual data, making them increasingly integral for generating features in various prediction tasks. By transforming product descriptions, ingredient lists, and other textual attributes into context-rich embeddings, LLMs can capture linguistic nuances that simpler text processing methods may overlook or that tabular data may fail to capture. When integrated with ML algorithms, LLM-based features can facilitate scalable assessments of food processing levels and enhance classification reliability for large datasets.

Real-world datasets containing food and nutrition-related data are notorious for incomplete entries, from incomplete ingredient lists to partial or missing nutrient values. LLM-based approaches are a way to deal with such missingness. Because an LLM processes input as a sequence, one can simply omit an unavailable field or include a placeholder (e.g., "unknown") without breaking the model's input format[108]. The model's contextual embedding will naturally reflect the absence of that information and down-weight its importance, focusing on the available data. Recent research on heterogeneous data imputation leverages this property by inserting mask tokens for missing entries and letting the language model's context understanding fill in or ignore gaps. In essence, the contextual representation produced by an LLM can capture what is known about a product (e.g., certain ingredients, a few nutrient values) while gracefully handling unknown portions. This starkly contrasts with traditional pipelines that might require explicit imputation, such as filling in missing nutrient values with averages or minima, which may not only introduce bias and errors but also require expertise and additional work.

By supporting variable-length and flexible inputs, LLM-based classifiers maintain performance even as data quality varies. In practice, this means we no longer must discard or heavily preprocess records with missing values; the model utilizes whatever information is present, making it highly practical for real-world, heterogeneous databases. This is particularly useful in the domain of food processing classification, where both descriptive (ingredient lists, additives, preparation techniques) and quantitative data (e.g., nutrient values) contribute to determining the level of processing. For example, FoodProX relies on a fixed panel of nutrients to infer processing scores; however, these models may underperform when significant nutrient data is missing and do not currently leverage the noisy but informative textual data captured by branded food product ingredient lists. In contrast, LLM-based models can integrate any available information, whether it is a complete ingredient list, partial nutrient panel, or descriptive metadata, to assess processing level with greater flexibility and less disambiguation efforts. As such, LLMs provide a robust framework for predicting NOVA classes or estimating food processing scores even under real-world data constraints, making them particularly valuable for scalable, automated food classification efforts. In the following section, we present a case study that sets up, trains, and compares machine learning pipelines based on FoodProX and LLM architectures. The primary goal of this work is to evaluate the emergent capabilities of pretrained LLMs when applied to nutritional data, rather than developing them from scratch.

## 12.5 Case Study: Applying Machine Learning and Large Language Models for Food Processing Classification

Since the early development of FoodProX, more datasets with metadata on food processing have become available. In this case study, we will utilize data from the Open Food Facts platform to train machine learning models and LLMs for predicting NOVA classes by combining structured and unstructured data. Open Food Facts is an open, crowd-sourced database of branded food products from around the world, based in France, mainly created through the efforts of thousands of volunteers systematically scanning foods in grocery stores[109]. The database includes a diverse set of attributes such as product names, ingredient lists, nutritional composition, food processing classifications (including NOVA groups, Nutri-Score), brand and packaging details, environmental impact indicators (e.g., carbon footprint), and food safety information (e.g., allergens and additives). The database is maintained by contributors who upload product details, making it a valuable resource for large-scale food analysis. While the quality control cannot compare to curated epidemiological datasets, Open Food Facts is increasingly used in research related to food processing, nutrition, and health due to its broad coverage and organized structure format[110]. Given its open-access nature, Open Food Facts provides an excellent foundation for machine learning and deep learning applications that aim to assess food processing characteristics, making it an ideal dataset for developing AI-driven models to classify foods by processing levels.

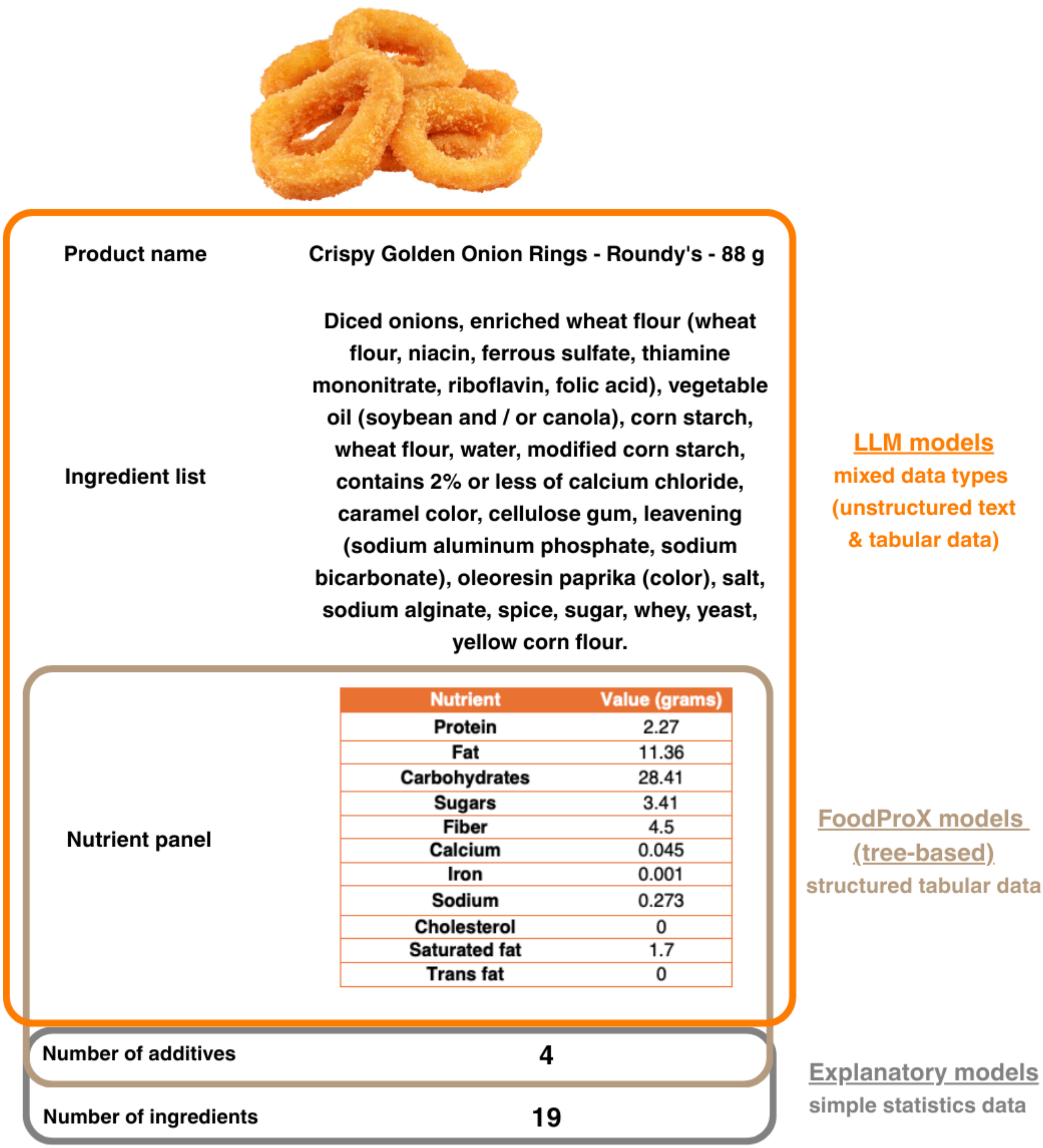

*Figure 3:* ***Example instance from the Open Food Facts dataset used in the training of the predictive models.*** *The product name, ingredient list, and full nutrient panel (not fully shown here) are used to construct the input sentences for the LLM-based models. The nutrient panel shown includes the 11 nutrients used to train the FoodProX-based models. The last two quantitative indicators are used in the explanatory models. The number of additives is also included as an additional feature in one variant of the FoodProX model.*

The metadata in the Open Food Facts dataset is heterogeneous, comprising both unstructured textual data (such as product descriptions and ingredient lists) and structured numerical data (including nutrient values commonly referred to as nutrition facts). We filtered the Open Food Facts dataset to include only products with English names and complete information for the key fields required in our analysis — product name, ingredient list, NOVA classification, and the 11 nutrients used in the FoodProX models — resulting in a final dataset of 149,960 products. To enhance interpretability and offer practical heuristics, we also incorporated simple engineered features extracted from this information (such as the total number of ingredients and the total number of additives). These variables are hypothesized to correlate with the likelihood that a product is classified as ultra-processed, offering simple, interpretable benchmarks to complement more complex predictive models. Figure 3 illustrates a representative example of the rich and diverse metadata available for a single food product in Open Food Facts. The figure is organized into distinct sections, each showcasing a different type of data that contributes to model development and analysis:

- Engineered features, such as the number of ingredients and the number of additives, provide a simple and interpretable metric for assessing food processing. These features act as rule-of-thumb indicators[111], where a higher count suggests that a food product is ultra-processed.

- Tabular numerical data, which includes detailed nutrient values (e.g., fats, carbohydrates, proteins, etc.), offers a precise and structured representation of a product's nutritional composition. This data is well-suited for traditional machine learning models, enabling rigorous analysis and comparison based on standardized measurements, although limited by missing values.

- Unstructured textual data, including food descriptions and ingredient lists, is processed using natural language techniques. By converting this information into semantic embeddings, LLM models capture the nuances and context of each food product, even when some information might be missing.

Together, these diverse data types offer complementary insights into the composition and quality of food products, enabling the development of a robust array of AI models for classifying levels of food processing that vary in complexity and interpretability.

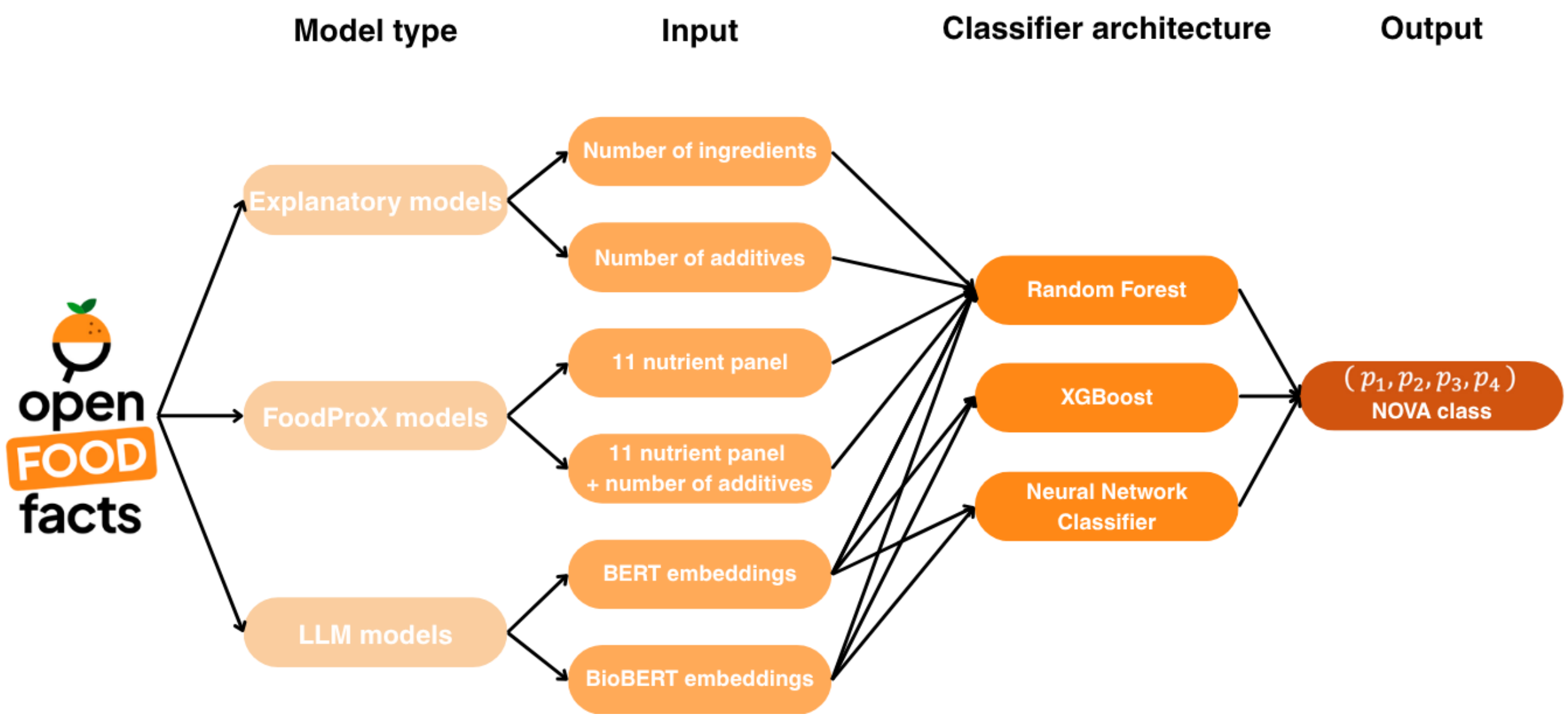

*Figure 4:* ***Case Study Schematic.*** *A diagram illustrating models, input data, and architecture types used in this study. All models are assessed by their ability to predict the NOVA class.*

## 12.5.1 Explanatory Models

Before examining more advanced classification approaches, we first established a baseline using two explanatory models to predict NOVA that rely on simple, yet informative features: the number of ingredients and the number of additives. In the ingredient-based model, each food is classified by how many individual components appear on its label, while in the additive-based model, classification is guided by the presence and quantity of listed additives. Both trained classifiers are Random Forest, and we use predefined model splits for cross validation, which remain consistent across the models (explanatory, FoodProX, and LLM-based models). Although these features alone cannot capture the full nuance of food processing, they offer insight into how much transformation or industrial intervention a product may have undergone. By focusing on these straightforward descriptors, we obtain a useful starting point for comparing and contextualizing the more complex FoodProX and LLM-based models introduced later in the chapter.

The explanatory models are trained using a Random Forest classifier with predefined splits and a five-fold cross-validation. Hyperparameter tuning is performed by grid search.

## 12.5.2 FoodProX Models

The second part of this case study involves applying the FoodProX algorithm to the Open Food Facts dataset, using two predictive models to classify foods according to the NOVA classification system. The first model uses the set of 11 nutrients as input features to predict the NOVA class of a food product. This approach leverages 11 key nutrients that provide a rich representation of each product's nutritional composition: proteins, fat, carbohydrates, sugars, fiber, calcium, iron, sodium, cholesterol, saturated fat, and trans-fat. The second model extends the first by including an additional input variable: the total number of additives in a food product. Alongside the original 11 nutrients features, this additive count aims to enhance the model's ability to estimate NOVA classifications. This inclusion aligns with criteria often used by manual assessors, who explicitly consider additives when identifying ultra-processed foods.

Like the explanatory models, both FoodProX models also use a Random Forest classifier, trained using the predefined splits and a five-fold cross-validation with a grid search for hyperparameter tuning.

### 12.5.3 Leveraging Large Language Models

In the final part of the case study, we train a prediction model using contextual vector representations of food products. To do this, we leverage pre-trained transformer-based models such as BERT (Bidirectional Encoder Representations from Transformers)[112] and its domain-specific variant, BioBERT[113]. These models have demonstrated strong performance across various biomedical and clinical tasks, including relation extraction from clinical texts[114], detection of adverse drug reactions from biomedical and social media sources[115], and identification of relationships between biomedical entities in scientific literature[116].

BioBERT is a domain-specific adaptation of BERT, pre-trained on large-scale biomedical corpora such as PubMed abstracts and PMC full-text articles. It is particularly well-suited for tasks involving technical or health-related terminology. In the context of food classification, BioBERT is expected to better capture the semantic meaning of ingredients and additives such as "lecithin," "ascorbic acid," or "monosodium glutamate," which appear more frequently in biomedical literature than in general-language corpora. By comparing models using BERT and BioBERT embeddings, we aim to assess whether this domain adaptation improves classification performance for food processing prediction.

These models provide context-aware embeddings that capture semantic relationships between words. By leveraging text embeddings as structured features, researchers have successfully used LLMs for food recommendation systems, ingredient substitution models[117], and dietary assessment automation[118]. We employ BERT and BioBERT to generate embeddings from Open Food Facts data, integrating food descriptions, ingredient lists, and nutrient values to classify foods according to NOVA classes. The process involves transforming product metadata into structured text that can be analyzed by the model. Specifically, we first create sentences from the available information for each food:

***"[FOOD NAME] has the ingredients: [INGREDIENT LIST], and the nutrients: [X UNIT OF NUTRIENT N1], [Y UNIT OF NUTRIENT N2], …"***

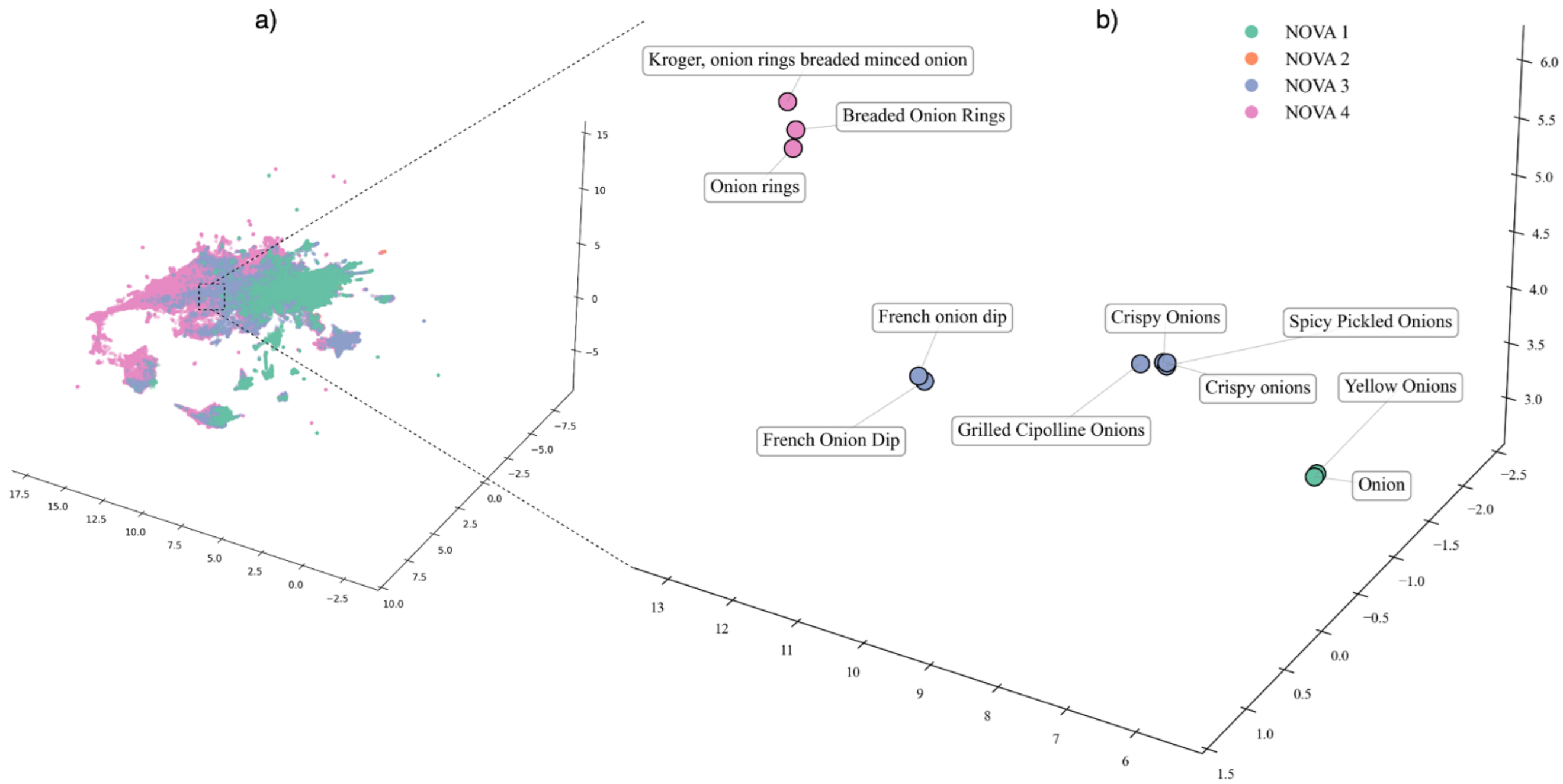

*Figure 5: a)* ***Three-dimensional UMAP projection of BERT embeddings colored by NOVA classification.*** *Each point represents a food item embedded using BERT and reduced to three dimensions using the Uniform Manifold Approximation and Projection (UMAP). Points are colored according to their NOVA group. This visualization illustrates the clustering of food items based on linguistic features in their names, showing the separation across NOVA categories. b) 3D UMAP projection of BERT embeddings for onion-based products, colored by NOVA classification. The sequence spans from raw foods (e.g., Onion, Yellow Onions which are NOVA 1) through intermediate forms (e.g., Grilled Cipolline Onions, French Onion Dip, and Crispy Onions which are NOVA 3) to ultra-processed items (e.g., Breaded Onion Rings from NOVA 4), illustrating how embedding positions shift across NOVA groups.*

For example: ***"Chocolate chip cookies have the ingredients: wheat flour, sugar, cocoa butter, chocolate liquor, and the nutrients: 50g of carbohydrates, 5g of protein, 10g of fat, 2g of fiber."***

By structuring the input in this manner, we enable the model to interpret both categorical and numerical data within a unified framework. Each sentence is then tokenized and passed through BERT/BioBERT, and rather than applying mean pooling across all token embeddings — which can dilute sentence-level meaning when sequences are long or contain uninformative tokens (e.g., units, numbers) — we extract the embedding corresponding to the [CLS] token. This 768-dimensional vector is specifically trained during BERT's pretraining to represent the entire input sequence, making it well-suited for downstream classification tasks where a holistic understanding of the input is required. It is important to note, however, that transformer-based language models are not inherently numerically grounded: even when nutrient quantities are included in the input text, models such as BERT treat numbers as tokens rather than true numerical magnitudes and therefore do not encode quantitative relationships (e.g., "50 g" is not inherently treated as ten times "5 g"). This limitation also applies to prompt-based approaches in which researchers manually query LLMs, such as ChatGPT, to assign NOVA labels from ingredient lists, nutrient values, and/or product descriptions. Prompting relies on the same token-level representations and may inconsistently weight quantitative information relative to structured models.

To explore how the semantic structure of the created sentences reflects the processing levels, we projected the generated BERT embeddings into three dimensions using the Uniform Manifold Approximation and Projection (UMAP). Figure 5a shows the global embedding landscape of the Open Food Facts database, with each point representing a food item colored by its NOVA classification. While the categories show partial overlaps, distinct regions emerge, particularly between unprocessed foods (NOVA 1) and ultra-processed foods (NOVA 4), indicating that the information contained in the food description, ingredient lists, and nutrient profiles, carries implicit information about processing level. Figure 5b zooms in on a subset of onion-based foods to illustrate a semantic trajectory from minimally processed (Onion, Yellow Onions) to ultra-processed products (Breaded Onion Rings, Kroger onion rings breaded minced onion). As processing increases, the placement of the embeddings shifts in the embedding space, suggesting that BERT captures subtle linguistic and conceptual transformations linked to processing intensity. Together, these visualizations demonstrate the potential of language models to encode meaningful processing-related patterns in food descriptions.

The [CLS] token embeddings extracted from each input sequence are then used as feature vectors to train a classification model that predicts NOVA classes. At this stage, one common approach is to fine-tune the BERT model, starting from its pretrained weights and continuing training to optimize the embeddings for the selected classification task. However, in our case, we did not fine-tune either BERT or BioBERT. Instead, we used the [CLS] embeddings as fixed input features for separate downstream classifiers. This decision was based on the strong classification performance already achieved and the fact that fine-tuning introduces substantially higher computational costs.

We trained three different classification architectures: two tree-based models (Random Forest and XGBoost) and one neural network. Each classifier was trained twice, once using the [CLS] embeddings from BERT and once using those from BioBERT. All LLM-based models were trained using the predefined data splits, with randomized hyperparameter tuning and five-fold cross-validation, in line with the training protocols adopted for both the FoodProX and explanatory models.

## 12.5.4 Comparing the Models

All models follow the same two-stage validation protocol. First, a stratified 20% of the dataset is held out for hyperparameter tuning. The remaining 80% is partitioned into five fixed train/test stratified folds for cross-validation. For the FoodProX and explanatory models, we perform a grid search over the hyperparameter space. In contrast, for the BERT and BioBERT models, we apply a randomized hyperparameter search to reduce computation time. Once optimal hyperparameters are selected using the tuning set, we train five independent models, one per fold, using these settings and report the average performance across the five held-out test sets. This standardized procedure ensures that all models are evaluated under identical data splits, allowing for fair and reproducible comparisons of classification performance (in terms of AUC, AUP, and related metrics).

The AUC and AUP values vary substantially across models and NOVA classes (Figure 6). Explanatory models based solely on ingredient and additive counts consistently underperform relative to other approaches, particularly for NOVA 2 and NOVA 3, where their predictive power is markedly limited. In contrast, both FoodProX models demonstrate strong discrimination (AUC) and precision–recall performance (AUP), confirming the high predictive value of nutrient composition when assessed holistically. The model using only 11 nutrients already achieves high separability across classes (AUCs: 0.988, 0.983, 0.926, 0.948; AUPs: 0.941, 0.815, 0.802, 0.974). Augmenting this with the additive count further improves performance, reaching AUCs of 0.993, 0.988, 0.966, and 0.980 and AUPs of 0.956, 0.860, 0.988, and 0.991. This agrees with the preliminary observations on Open Food Facts reported in [94].

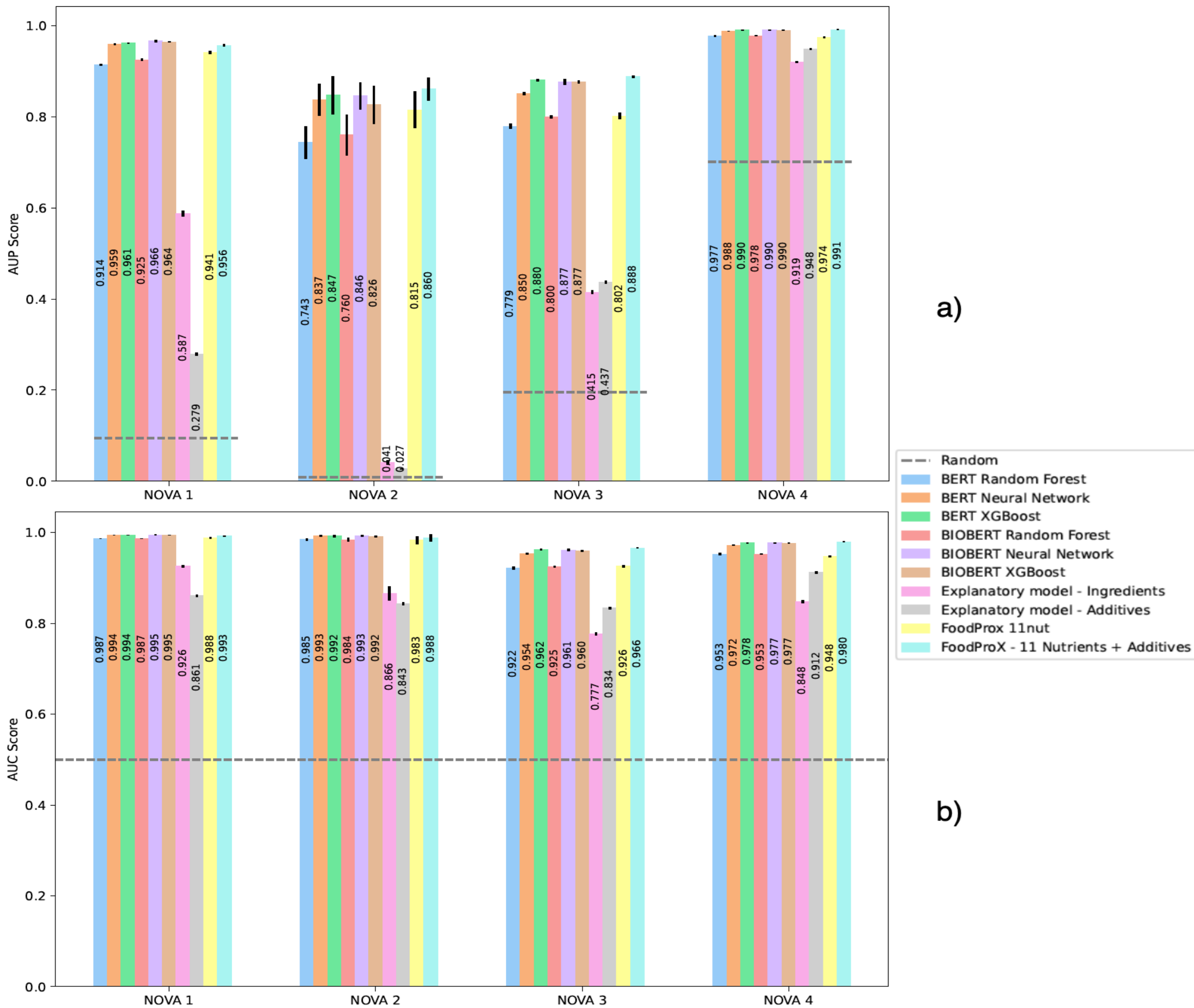

*Figure 6:* ***Comparative AUP and AUC scores for each classification model across the four NOVA classes (1–4), illustrating precision–recall trade-off (AUP) and model discrimination (AUC) per class.*** *a) The Precision–Recall curve plots precision (positive predictive value) versus recall (sensitivity) over different thresholds, particularly informative on imbalanced datasets. Area Under the Precision–Recall Curve (AUP) summarizes the balance between capturing true positives (recall) and limiting false positives (precision). A high AUP indicates the model maintains both high precision and high recall, especially important when class prevalence varies widely (as with NOVA classes). b) The ROC (Receiver Operating Characteristic) curve plots the true positive rate (sensitivity) against the false positive rate (1 – specificity) at varying classification thresholds. AUC (Area Under the ROC Curve) quantifies overall model discrimination: an AUC of 0.5 indicates random performance, while 1.0 denotes perfect class separation. Higher AUC values mean the model better distinguishes between positive and negative instances, regardless of threshold choice.*

BERT- and BioBERT-based classifiers achieve among the highest AUCs overall — up to 0.995 for NOVA 1 and 0.993 for NOVA 2 — demonstrating their capacity for capturing rich contextual patterns. However, their AUP scores, while strong, are more comparable to other models except for NOVA 1 (AUP = 0.966). This suggests that while LLM-derived embeddings excel in overall ranking and discrimination, they yield broader probability distributions for ambiguous items, which can reduce precision–recall balance at fixed thresholds. Additionally, the use of randomized hyperparameter search for these models, due to computational constraints, may have limited their ability to fully match the grid-searched FoodProX models.

### *Class-by-Class Observations*

- **NOVA 1 (Unprocessed/Minimally Processed)**
  All advanced classifiers demonstrate excellent performance in identifying unprocessed or minimally processed foods. The simplest baseline models, based solely on ingredient or additive counts, perform the worst, with the ingredient count model outperforming the additive-based one. Among the more sophisticated methods, the BERT/BioBERT models paired with neural networks and XGBoost achieve the highest AUC scores (0.994 and 0.995), followed closely by FoodProX with 11 nutrients plus additive count (AUC = 0.993) and FoodProX using only 11 nutrients (AUC = 0.988). In terms of precision–recall performance, the embedding models again lead with AUP scores of 0.966 and 0.964, respectively, followed by FoodProX 11 nutrients plus additives (AUP = 0.956) and FoodProX 11 nutrients (AUP = 0.941). Overall, nutrient-based models alone already achieve near-perfect separability for NOVA 1. Incorporating contextual embeddings from text provides modest but consistent improvements in both ranking (AUC) and precision–recall (AUP).

- **NOVA 2 (Processed Culinary Ingredients)**
  NOVA 2 is the smallest and most challenging category to classify. The BERT and BioBERT neural network classifiers achieve the highest AUC scores at 0.993, followed by the XGBoost models at 0.992. FoodProX model with 11 nutrients plus additive count reaches AUC = 0.988, while the simpler FoodProX 11-nutrient model scores AUC = 0.983. When comparing AUP scores, the FoodProX 11 nutrients plus additives model leads with 0.860, closely followed by the BERT XGBoost model (0.847) and the BioBERT neural network (0.846). The FoodProX model without additive count scores 0.815, while the BERT and BioBERT random forest classifiers record the lowest AUPs among the embedding models (0.743 and 0.760, respectively). FoodProX's advantage lies in producing well-separated probability estimates for NOVA 2, allowing for a clearer threshold that balances precision and recall, thus improving AUP. By contrast, although embedding-based models slightly outperform in overall ranking ability (as reflected in higher AUC), their probability distributions for NOVA 2 tend to overlap more with those of other classes. This may stem from the ambiguous role of culinary ingredients, which often appear as ingredients in other NOVA classes. As a result, it becomes more challenging for these models to identify a single decision threshold that preserves both high precision and high recall.

- **NOVA 3 (Processed Foods)**
NOVA 3, like NOVA 2, is among the most challenging classes to classify. This is due primarily to the variability of food products within the category and the inconsistency in manual labeling. The highest AUC of 0.966 is achieved by the FoodProX model using 11 nutrients combined with additive count, followed closely by the BERT XGBoost model at 0.962 and the BioBERT Neural Network at 0.961. In terms of AUP, FoodProX 11 nutrients plus additives again leads with 0.888, followed by BERT XGBoost at 0.880. The FoodProX model with only 11 nutrients performs reasonably well (AUC = 0.926, AUP = 0.802), while the simpler ingredient and additive count baselines perform substantially worse (ingredient count AUC = 0.777, AUP = 0.415; additive count AUC = 0.834, AUP = 0.437). Their improvement over NOVA 2 results reflects the fact that NOVA 3 foods often contain multiple ingredients and additives, making these basic features somewhat more informative and variable. These results show that supplementing the nutrient panel with a count of additives increases FoodProX's performance and brings it in line with the top-performing embedding models.

- **NOVA 4 (Ultra-Processed Foods)**
As the majority class, NOVA 4 is the easiest to classify across all models, as these products typically list multiple additives and have distinctive nutrient profiles. Even the simplest exploratory models perform strongly in this category, achieving an AUP of 0.919 with the ingredient count and 0.948 with the additive count. The FoodProX model using only 11 nutrients attains an AUC of 0.948 and an AUP of 0.974. When the additive count is included, these metrics increase to an AUC of 0.980 and an AUP of 0.991. Embedding-based models perform comparably, with top AUCs reaching 0.978 and AUPs up to 0.990. In this class, both the enhanced FoodProX model and the leading embedding pipelines reach top-level performance, indicating that either nutrient composition or contextual embeddings are sufficient to reliably classify ultra-processed foods.

### *Time and Computational Resources*

Model runtimes varied widely depending on input dimensionality and algorithmic complexity. The two explanatory models completed hyperparameter tuning in about 2 minutes and the full five-fold cross-validation in under 3 minutes. The FoodProX Random Forest model on 11 nutrients required around 22 minutes of grid search hyperparameter tuning plus 8 minutes of cross-validation model training, totaling 30 minutes, while when additives were included, the total runtime remained under 30 minutes. In contrast, the embedding-based classifiers ran substantially longer: the neural network classifiers built on BERT or BioBERT embeddings took 50-75 minutes end-to-end (approximately 55 minutes for BioBERT, and approximately 1 hour and 14 min for BERT), while the tree-based models using BERT and BioBERT embeddings as features required multiple hours (e.g., two hours and 34 min for BioBERT Random Forest classifier and over 3 hours for BioBERT XGBoost classifier). All experiments were carried out on a CPU-only server (36 physical cores, 64 GB RAM). Generating the embeddings on the sentences took about 3 hours using 36-core CPU with 64 GB of RAM, but less than 15 minutes when using 4 Nvidia V100 SXM2 GPU. Hyperparameter search ranges were chosen to balance computational tractability and predictive performance, and, as our results show, these constraints had minimal impact on final accuracy.

### *Model Benefits and Practical Implications*

FoodProX marks a substantial step forward in food processing classification by addressing critical limitations of simpler, manually curated systems:

- **Scalability:** Its automated, nutrient-based framework enables high-throughput classification across large food databases with modest computational resources.
- **Reproducibility:** The model relies on standardized and well-regulated nutrient data, ensuring consistency across datasets.

LLM-based models, such as those built on BERT and BioBERT, offer complementary advantages that make them particularly effective in handling branded food products:

- **Context-Aware Representation:** The [CLS] token captures high-level interactions between ingredients and nutrients, improving classification accuracy compared to averaging token embeddings.
- **Robustness to Missing Data and Input Heterogeneity:** The contextual associations enable the models to infer patterns even when faced with incomplete nutritional fields and poor standardization of the ingredients and their synonyms.
- **Automation & Efficiency:** LLM-based methods reduce manual feature engineering efforts, making predictive analysis more efficient and scalable.

LLM-based models are particularly well-suited for real-world applications. Their resilience to inconsistent or incomplete records, common in large-scale datasets like Open Food Facts, makes them reliable for deployment in practice. Moreover, their flexibility allows for seamless incorporation of new information without requiring architectural adjustments. Pretrained LLMs can be fine-tuned on large volumes of food data, adapted to multilingual data, and scaled globally, reducing the need for extensive manual normalization.

Importantly, all model types discussed in this study, whether based on nutrient composition, additive counts, or contextual embeddings, can also be used to compute the continuous FPro score. Since FPro is derived from the model's class probability vector, it is agnostic to the underlying architecture. Regardless of whether the classifier is a FoodProX model or an LLM-based approach, FPro can be calculated as the projection of the predicted class probabilities onto the axis spanning from NOVA 1 to NOVA 4. This flexibility allows practitioners to select a modeling strategy best suited to their data and computational constraints, while still accessing both discrete NOVA labels and a nuanced, continuous measure of food processing. Although this study has focused primarily on NOVA classification performance, the same outputs can readily support FPro-based analyses without any modification to the models.

## *Limitations and Future Perspectives*

While the results presented here are encouraging, several limitations must be acknowledged, along with opportunities for future refinement. First, the quality and consistency of the underlying data remain a significant constraint. Although the Open Food Facts dataset is large, diverse, and publicly accessible, it contains inconsistencies that can affect model reliability. For example, formatting errors in nutrient values, such as the use of decimal commas instead of points, occasionally led to implausibly high concentrations of micronutrients like vitamin A and vitamin C. Additionally, some NOVA labels appear to be misassigned, with foods more appropriately classified as NOVA 3 being labeled as NOVA 2. Such noise introduces uncertainty into both training and evaluation, and may account for some misclassifications across models.

Second, the dataset suffers from class imbalance, with NOVA 2 being substantially underrepresented. This makes it particularly difficult for models to learn reliable patterns for culinary ingredients, which often lack distinctive signals in either nutrient composition or additive profiles. To address this imbalance, future research could explore how to tailor resampling techniques, class weighting strategies, or data augmentation methods to the unique challenges of food composition data, while also accounting for its inherent redundancy.

Third, while the use of LLMs has proven effective, the construction of input text could be further optimized. Currently, inputs are created by concatenating product names, ingredient lists, and nutrition facts into a single sequence. A more structured representation may help the model better isolate which components carry the most predictive signal. Future work should therefore focus on improving interpretability, for example through feature-importance analyses to identify which phrases or descriptors contribute most to predictions, as well as experiments with alternative prompt structures or token-level attention mechanisms. Enhancing interpretability is particularly important given the trade-off between predictive performance and model transparency. Unlike models such as FoodProX, which provide interpretable outputs by quantifying the contribution of individual features (e.g., specific nutrients or additive counts), LLM-based models largely function as black boxes: their contextual embeddings make it difficult to trace predictions back to specific input elements. In addition, as discussed earlier, traditional transformer architectures do not inherently encode numerical magnitude when nutrient values are represented as text, suggesting that future approaches may benefit from hybrid embeddings that combine contextual language representations with explicitly modeled numerical features.

Finally, this work underscores the potential of automated food classification systems for public health monitoring and research, particularly when applied to large-scale datasets. Across all NOVA classes, combining nutrient profiles with either textual embeddings (as in BERT and BioBERT models) or simpler features like additive counts (as in FoodProX) consistently outperformed models that rely on single, coarse-grained features, often leveraged by manual assessors. For projects with limited computational resources, FoodProX models offer a strong balance of efficiency and accuracy. In more resource-rich settings, LLM-based embeddings may provide added value, particularly when dealing with ambiguous or borderline food items, by leveraging the deeper context contained in product descriptions and ingredient lists. As food environments become increasingly digitalized, integrating fine-grained food processing information into emerging AI models will further enhance our ability to capture the complex properties of foods, from the structure of the food matrix and its transformations during processing to the broader exposures shaping dietary patterns. Coupled with the rapidly increasing dimensionality of personal data generated by dietary tracking technologies and other digital measures of behavior and consumption, such approaches may ultimately reveal emergent relationships between food processing, diet, and health, enabling more precise and data-driven assessments of dietary quality and disease risk.

# Code Availability

All codes generated for the analysis are available through Dr. Menichetti's Lab GitHub repository at https://github.com/menicgiulia/AI4FoodProcessing.

# Acknowledgments

G.M. is supported by NIH/NHLBI grant K25HL173665 and 24MERIT 1185447. This manuscript is a preprint version of Chapter 12 in "Agrifood Informatics" (The Royal Society of Chemistry, 2026).

# Glossary

1. **Additives.** Compounds intentionally added to food to extend shelf life, enhance flavor, alter appearance, or restore lost nutrients.
2. **Area Under the Receiver Operating Characteristic Curve (AUC).** A metric for evaluating classifier performance that measures a model's ability to distinguish between classes across all possible decision thresholds. A value of 1.0 indicates perfect discrimination; 0.5 indicates no better than chance.
3. **Area Under the Precision-Recall Curve (AUP).** A classification performance metric that captures the trade-off between precision (fraction of positive predictions that are correct) and recall (fraction of true positives identified). AUP is particularly informative when class distributions are imbalanced.
4. **Bidirectional Encoder Representations from Transformers (BERT).** A general-purpose transformer-based large language model used to generate contextual text embeddings from sequences of text.
5. **BioBERT.** A domain-specific adaptation of BERT further pre-trained on large biomedical corpora, including PubMed abstracts and PubMed Central full-text articles.
6. **Deformulation.** A reverse-engineering approach in which ingredient lists are analyzed to infer the processing steps used to produce a food.
7. **Embeddings.** Dense numerical vector representations of text generated by large language models. Unlike simple word counts, embeddings capture the semantic meaning and relationships between words.
8. **Food Matrix.** The complex physical and chemical organization of nutrients and bioactive compounds within a whole food, which governs digestion, absorption, and satiety.
9. **Food Processing.** Any deliberate intervention applied to a raw agricultural product that alters its form, flavor, shelf life, or safety.
10. **FoodEx2.** A hierarchical food classification system developed by the European Food Safety Authority for food safety monitoring and dietary exposure assessment.
11. **FoodOn.** A comprehensive food ontology providing standardized, hierarchical vocabulary for describing food products, ingredients, and transformation processes.
12. **FoodProX.** A machine learning classifier that predicts the degree of food processing from a food's full nutrient composition and trained on NOVA-labeled data.
13. **FPro.** A continuous score ranging from 0 (minimally processed) to 1 (ultra-processed), derived from a food’s NOVA class probability vector.
14. **Front-of-Pack (FOP) Labeling.** A consumer-oriented labeling approach displaying a simplified nutritional and processing summary prominently on food packaging sold at retailers.
15. **Highly Processed Foods (HPFs).** A broad term for multi-ingredient, industrially formulated food products that are typically ready to heat or eat and include components uncommon in traditional cooking.
16. **Hyper-palatability.** A property of foods engineered to maximize sensory appeal by combining fat, salt, and sugar or refined carbohydrates at levels rarely found in unprocessed foods.
17. **LanguaL.** A multilingual, hierarchical food indexing thesaurus used to standardize food descriptions across databases and research applications.
18. **Large Language Model (LLM).** A deep learning model trained on large text corpora to

learn statistical patterns in language, enabling more flexible handling of variable or incomplete input data.

19. **Markers of Ultra-Processing (MUPs).** Ingredients or additives derived from industrial processes that indicate ultra-processed food production as defined under the SIGA food classification system.
20. **Minimally Processed Foods (MPFs).** Foods belonging to NOVA 1 that have undergone only minor interventions such as cleaning, portioning, freezing, boiling, or pasteurization that do not substantially transform the food or its nutritional structure.
21. **NOVA.** A four-group system organizing foods by the extent and purpose of processing: NOVA 1 (unprocessed or minimally processed foods), NOVA 2 (processed culinary ingredients), NOVA 3 (processed foods), and NOVA 4 (ultra-processed foods).
22. **Nutri-Score.** A front-of-pack labeling system assigning a letter grade from A to E based on the weights of negative components (energy, sugars, saturated fat, sodium) against positive components (fiber, protein, fruit and vegetable content).
23. **Ontology.** A structured, hierarchical vocabulary that defines standardized terms and their relationships, enabling consistent classification and integration of data across platforms.
24. **Processed Culinary Ingredients (PCIs).** Foods belonging to NOVA 2, derived from natural sources and intended for use in cooking rather than direct consumption such as oils, flour, and butter.
25. **Random Forest.** An ensemble machine learning algorithm that builds multiple decision trees and combines their outputs for classification or regression.
26. **Ready-to-Eat (RTE).** A term describing food products that require no further cooking or preparation before consumption.
27. **SIGA.** A seven-class extension of the NOVA framework that further subdivides processing categories based on nutritional balance of fat, sugar, and salt as well as the presence of markers of ultra-processing.
28. **Ultra-Processed Foods (UPFs).** Foods in NOVA 4, industrially formulated products made predominantly from processed culinary ingredients and additives, with little or no intact whole food.
29. **Uniform Manifold Approximation and Projection (UMAP).** A dimensionality reduction technique used to project high-dimensional data, such as text embeddings, into two or three dimensions for visualization.